\pdfoutput=1

\documentclass[11pt]{article}

\usepackage[final]{acl}

\usepackage{times}
\usepackage{latexsym}

\usepackage[T1]{fontenc}

\usepackage[utf8]{inputenc}

\usepackage{microtype}
\usepackage{xspace}

\usepackage{microtype}
\usepackage{hyperref}
\usepackage{url}
\usepackage{booktabs}
\usepackage{graphicx}
\usepackage{caption}
\usepackage{subcaption}
\usepackage{multirow}
\usepackage{cleveref}
\usepackage{xspace}
\usepackage{verbatim}
\usepackage{listings}
\usepackage{adjustbox}
\usepackage{enumitem}
\usepackage[normalem]{ulem}
\useunder{\uline}{\ul}{}

\newcommand\todo[1]{\textcolor{purple}{[TODO: #1]}}

\usepackage{array}
\newcommand{\PreserveBackslash}[1]{\let\temp=\\#1\let\\=\temp}
\newcolumntype{C}[1]{>{\PreserveBackslash\centering}p{#1}}
\newcolumntype{R}[1]{>{\PreserveBackslash\raggedleft}p{#1}}
\newcolumntype{L}[1]{>{\raggedright\arraybackslash}p{#1}}

\definecolor{red}{RGB}{255, 0, 0}
\definecolor{lue}{RGB}{135, 206, 250}
\definecolor{green}{RGB}{205, 255, 204}

\newcommand{\datasetname}{ArxivDIGESTables\xspace}
\newcommand{\harddatasetname}{ArxivDIGESTables-Clean\xspace}
\newcommand{\silverdatasetname}{ArxivDIGESTables-Silver\xspace}
\newcommand{\baselinesname}{Task Decomposition\xspace}

\definecolor{lightblue}{rgb}{.8,.8,1}

\title{Intent-aware Schema Generation And Refinement\\For Literature Review Tables}

\author{Vishakh Padmakumar$^{1}$\thanks{Work done during summer internship at AI2, now at Stanford University}~~~~~Joseph Chee Chang$^{2}$~~~~~Kyle Lo$^{2}$\vspace{0.15cm}\\~~~~~\textbf{Doug Downey$^{2,3}$}~~~~~\textbf{Aakanksha Naik$^{2}$}\vspace{0.15cm}\\ 
 $^1$New York University~~~~~$^2$AI2~~~~~$^3$Northwestern University \vspace{0.15cm} \\
 {\tt\small vishakh@nyu.edu} \\
}

\begin{document}


\maketitle

\begin{abstract}
The increasing volume of academic literature makes it essential for researchers to organize, compare, and contrast collections of documents. Large language models (LLMs) can support this process by generating schemas defining shared aspects along which to compare papers. However, progress on schema generation has been slow due to: (i) ambiguity in reference-based evaluations, and (ii) lack of editing/refinement methods. Our work is the first to address both issues. First, we present an approach for augmenting unannotated table corpora with \emph{synthesized intents}, and apply it to create a dataset for studying schema generation conditioned on a given information need, thus reducing ambiguity. With this dataset, we show how incorporating table intents significantly improves baseline performance in reconstructing reference schemas. We start by comprehensively benchmarking several single-shot schema generation methods, including prompted LLM workflows and fine-tuned models, showing that smaller, open-weight models can be fine-tuned to be competitive with state-of-the-art prompted LLMs. Next, we propose several LLM-based schema refinement techniques and show that these can further improve schemas generated by these methods. 

\end{abstract}

\section{Introduction}
\label{sec:intro}

\begin{figure}
    \centering
    \includegraphics[width=0.88\columnwidth]{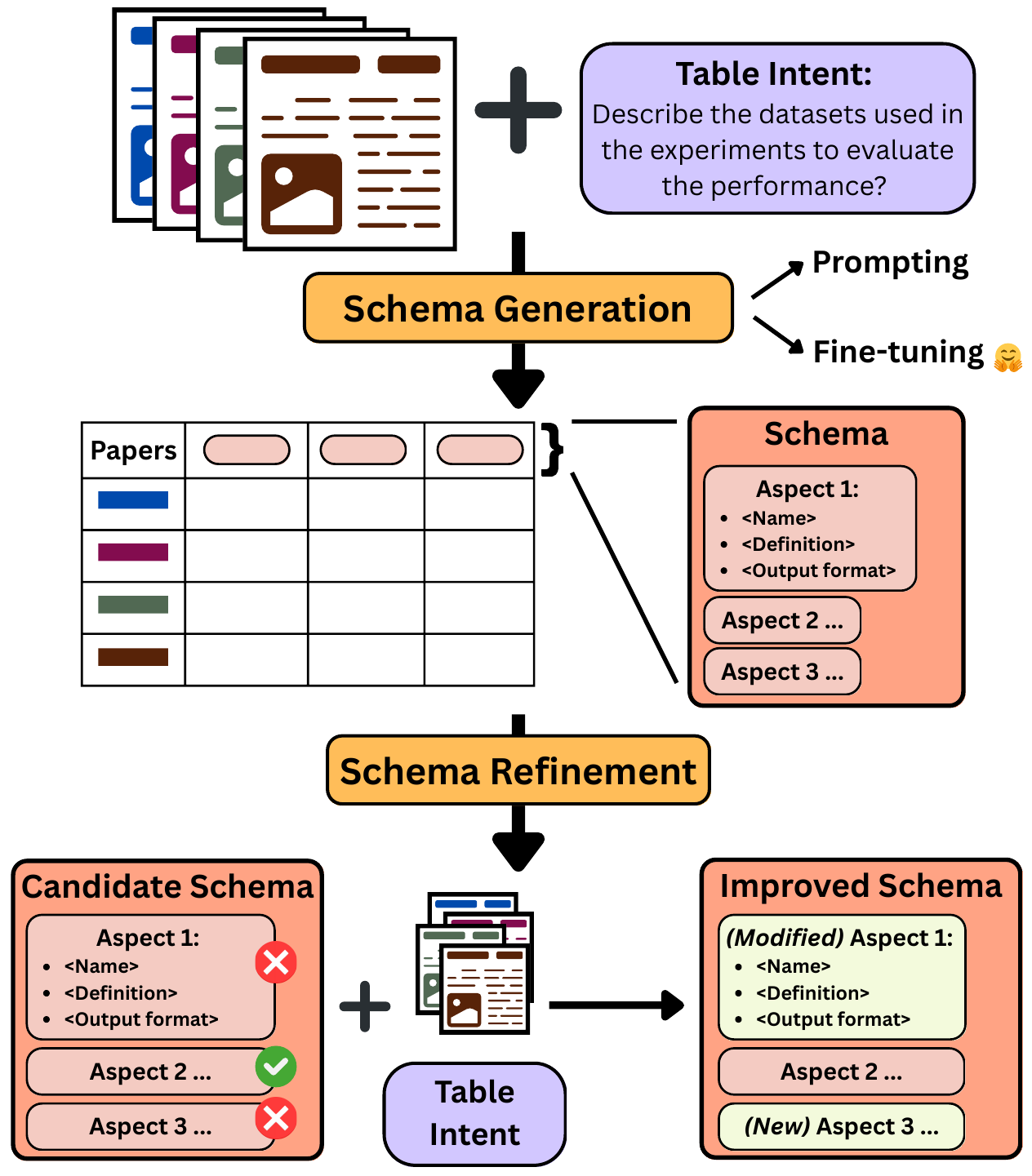}
    \caption{We benchmark the abilities of LLMs to perform schema generation for literature review tables given a table intent or information need (\S\ref{sec:dataset}) via prompting and fine-tuning (\S\ref{sec:generating_schemas}) as well as editing existing schema candidates (\S\ref{sec:editing_schemas}).
    \vspace{-0.5cm}
    }
    \label{fig:fig1}
\end{figure}
A common use case of large language models (LLMs) is synthesizing information from collections of documents
\citep{zhao2024wildchat, zheng2024lmsyschatm, openai2025deepresearch, google2024deepresearch}, a task that has become increasingly important in scientific domains amid the rapid growth of published literature
\citep{bornmann2021growth}. One important example of this task is the generation of literature review {\em tables} from sets of papers that compare them along shared aspects---a common target of literature review systems  \citep{elicit2024, singh2025ai2, fok2025facetstaxonomiessynthesesnavigating,wang2024scidasynth,Hashimoto2017AutomaticGO}. Typically, generating a literature review table involves two subtasks: (1) schema generation---identifying a relevant set of shared aspects to compare and contrast papers, which correspond to table’s columns, and (2) value generation---determining the content for each aspect–paper pair, i.e., filling in the table entries.
While value generation can be framed as question-answering, drawing on recent advances in retrieval-augmented generation (RAG) \citep{Lala2023PaperQARG, hilgert2024evaluating, singh2025ai2}, schema generation remains a more open problem without an obvious reduction to standard tasks. 

\begin{figure*}
    \centering
    \includegraphics[width=\linewidth]{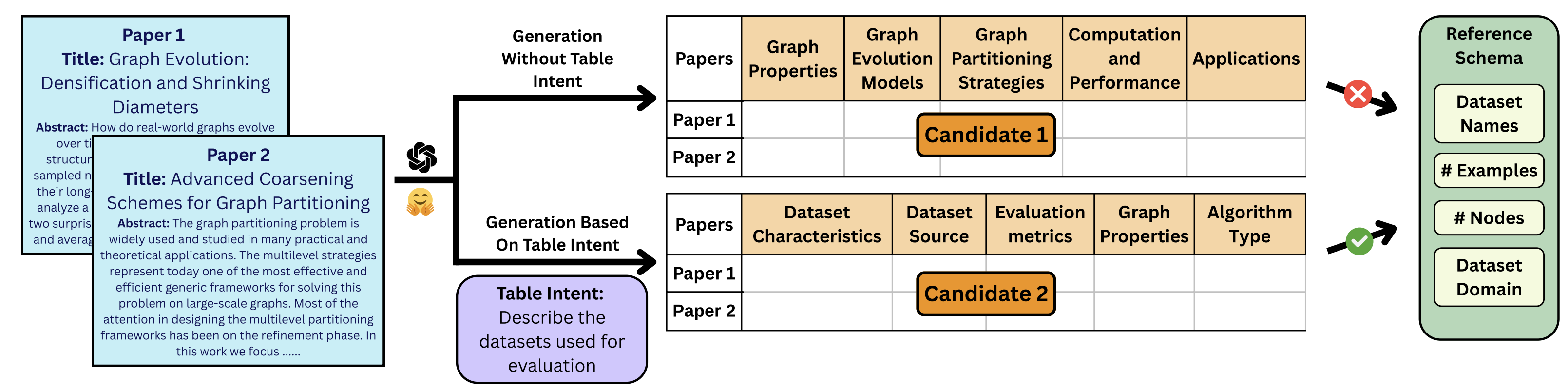}
    \caption{While \emph{Candidate 1} is a valid schema to compare the research papers, incorporating a table intent yields \emph{Candidate 2}, which better aligns with the user-written reference reflecting their actual information need. (\S\ref{sec:dataset}) 
    }
    \label{fig:goals_motivn}
\end{figure*}

Identifying appropriate dimensions along which to compare documents is a longstanding problem in information organization and sensemaking \citep{russell93cost}. Two issues have hampered progress on this problem: (1) ambiguity inherent in any reference-based evaluation process, and (2) lack of schema editing/refinement approaches given the open-ended nature of the task. Our work is the first to address both these issues.

Any reference-based evaluation for schema generation is inherently ambiguous because there can exist \emph{multiple valid schemas} for comparing the same set of papers. 
In fact, we argue the ideal choice of schema depends on the table's {\em intent}---the specific communicative goal or information need the author aims to address within the broader narrative of the paper, a notion missing from existing benchmarks \citep{newman2024arxivdigestables, hsu2024chime}. 
To illustrate this, consider Figure~\ref{fig:goals_motivn} where the papers being compared concern graph theory. \emph{Candidate 1} is a valid schema that selects aspects connected with the algorithms in both papers. However, this fails to match the actual table intent, which is meant to discuss the differences in the data used. Explicitly incorporating the table intent yields a schema much better aligned with this need, such as \emph{Candidate 2}. 

We address this gap by introducing an approach to augment unannotated table corpora with inferred intents in the form of open-ended research questions reflecting different aspects along which users might compare papers.  We demonstrate our approach on a recent table schema dataset, \datasetname  \citep{newman2024arxivdigestables}, 
showing that adding our intents as additional input to guide schema generation helps align model outputs with specific information needs, allowing for more principled evaluation 
and increased model performance (\S\ref{sec:dataset}). 
To address the lack of schema refinement methods and building on the observation that human authors often iteratively revise literature review tables, we evaluate a range of LLM-based techniques to improve tables by editing them  (\S\ref{sec:editing_schemas}). 

We first establish a strong set of single-shot generation methods by comprehensively benchmarking a broad range of prompting schemes for schema generation, varying both in how they use different levels of information from the papers being compared and in how they incorporate table intents into the pipeline \citep{lam2024concept, newman2024arxivdigestables, wang2025can}. 
Unlike prior work, we also fine-tuned open-weight models for schema generation (\S\ref{sec:training_for_schemas}) that are competitive to frontier LLMs already used in products, offering high value to the research community at a significantly lower inference cost. Then, we examine three categories of LLM-based techniques to refine existing schema candidates that support varying levels of user control:  Unguided editing, where a model is trained to revise schemas without explicit instruction, 
(\S\ref{sec:residual_editing}), 
Heuristic-guided editing, where models are fine-tuned to perform atomic operations like adding or removing aspects, 
(\S\ref{sec:heuristics_editing}),
Critique-guided editing, where models edit schemas by generating and applying natural language critiques 
(\S\ref{sec:critiques_editing}). However, generating reliable critiques is challenging, and that the small heuristic edits most consistently improves schema candidates (\S\ref{sec:editing_results}).  

In all, our contributions are: (1) An approach for augmenting unannotated table corpora with intents, (2) Benchmarking diverse schema generation methods (various prompting workflows vs finetuning, closed vs open models) demonstrating the value of table intents as well as identifying trade-offs in performance based on how various methods process papers, and (3) Proposing three approaches to editing table schemas that demonstrate ways to consistently improve existing candidates. We contribute a range of open-source artifacts, including \datasetname augmented with table intents, and a suite of open-weight LLMs trained for schema generation and editing to further research.\footnote{Open source data and models are listed in \Cref{sec:released_materials}; code is available at  \url{https://github.com/vishakhpk/arxivdigestables-with-intent}.}

\section{Inferring Table Intents for Better Task Specification}
\label{sec:dataset}

To assess LLM capabilities on schema generation, we evaluate their performance on producing literature review tables, a task recently formalized by \citet{newman2024arxivdigestables} in the \datasetname dataset. This dataset is a collection of $2,228$ literature review tables scraped from arXiv papers. Each instance consists of a table, where the rows represent each paper being compared and the columns represent the aspects of comparison. We treat the set of columns as the target schema.  

\subsection{Issues with current task specification}
\label{sec:motivation_goals}
\citet{newman2024arxivdigestables} observe that there are often multiple valid schemas which can be used to compare a fixed set of papers. In practice, authors typically construct tables to satisfy a specific information need or communicative goal within the narrative of the paper. We refer to this as the \emph{table intent}. Explicitly considering this intent is a natural way to disambiguate various candidates during schema generation. However, these intents are not usually recorded. \citet{newman2024arxivdigestables} use table captions and references to the table within the paper's text as proxies for table intent. But captions are written to be consumed alongside the table, and are often brief and incomplete representations of table intent (\Cref{fig:goals_motivn}). In-text references, on the other hand, tend to highlight specific information/trends from the table, potentially revealing some columns or values. In this work, we instead synthetically generate table intents and augment all examples in \datasetname.\footnote{We acknowledge contemporary work \citet{wang2025can} that also notes this limitation in the formulation of \citet{newman2024arxivdigestables} and detail the distinction to our work in \S\ref{sec:related}.}

\subsection{How we create table intents}
\label{sec:creating_goals}

We aim to construct \emph{table intents} that can guide schema generation by specifying the information need behind a literature review table. 
We synthetically generate these intents in the form of open-ended questions that the table is intended to answer, by prompting an LLM (GPT-4o) with the table content, caption, in-text references, and titles and abstracts of papers being compared. We provide the prompt used in \S\ref{sec:prompt_create_user_goals}, and hyperparameters for prompting in \S\ref{sec:appendix_user_goals}.
To curate high-quality intents, we generate five candidates per example and select the best one using an LLM-as-judge rubric provided in \S\ref{sec:prompt_select_user_goals}. We confirm the validity of the LLM judge selections with human evaluation in \S\ref{sec:appendix_user_goals}.

\subsection{Improving task specification and evaluation}
\label{sec:formulation_eval}
\paragraph{Task formulation} With our synthetically generated table intents, we define the {\em schema generation} task as follows: given a set of $M$ research papers $d_{1 \dots M}$ and a table intent $q$, generate a schema with $N$ aspects ($N \geq 2$), where each aspect corresponds to a column in a literature review table comparing the papers. Each aspect consists of: (1) the name, (2) a definition explaining its relevance for comparison, and (3) an output format describing how values for the column should be expressed (see \S\ref{sec:appendix_example_schema_json} for an example). We focus solely on the more challenging schema generation sub-task, differing from \citet{newman2024arxivdigestables}.

\paragraph{Evaluation} To match generated schemas to reference schemas for evaluation, we adopt the same schema alignment framework as \citet{newman2024arxivdigestables}. Each generated aspect is matched against each reference aspect using BERTScore \citep{Zhang2020BERTScore}, computed over the concatenated name, definition, and output format. An aspect pair is considered a match if the similarity exceeds a given threshold, which is a hyperparameter of the evaluation. To overcome the challenge of selecting the \emph{optimum} threshold, we compute precision, recall, and F1 over a range of thresholds and report the area under the curve (AUC), computed using the trapezoidal rule over all thresholds, for each metric. Specifically, we calculate metrics at each threshold value in increments of $0.01$ from $0.4$ to $1.00$ and report AUC within that range.\footnote{We select $0.4$ as the lower bound as our initial experiments showed that most schema candidates have a recall alignment of close to $1$ at that threshold so lower values do not provide distinguishing signal.} 

\subsection{Validating improvements in task specification}
We assess the utility of our synthetic table intents by evaluating whether they improve the performance of a baseline schema generation method. For these experiments, we adopt the schema generation approach from  \citet{newman2024arxivdigestables}, which prompts an LLM with information from all $M$ papers gathered into a single prompt. We evaluate schema generation in three settings: (i) where the input contains only  \emph{paper titles and abstracts} (\S\ref{sec:prompt_joint_title_abs}), (ii) where we add  \emph{table intents} in addition to \emph{paper titles and abstracts} (\S\ref{sec:prompt_title_abs_intent}), and (iii) where we alternatively add \emph{the table caption, and in-text references} as a means of directing schema generation (\S\ref{sec:prompt_cap_intext}). \S\ref{tab:intent_results} presents the results of these experiments using GPT-4o and Claude Sonnet-3.7 --- our table intents demonstrate clear utility, improving performance by $\sim5$ F1 AUC points on average across the three models while also outperforming captions and in-text references. 

\begin{table}[]
\resizebox{\columnwidth}{!}{%
\begin{tabular}{@{}rrccc@{}}
\toprule
\textbf{Model} & \textbf{Input} & \textbf{Recall} & \textbf{Precision} & \textbf{F1} \\ \midrule
\multirow{3}{*}{\textbf{GPT-4o}} & \textbf{T + A} & 0.1954 & 0.1791 & 0.1806 \\
 & \textbf{T + A + C + IR} & 0.2409 & 0.2159 & 0.2214 \\
 & \textbf{T + A + TI} & 0.2811 & 0.2648 & 0.2666 \\ \midrule
\multirow{3}{*}{\textbf{Sonnet-3.7}} & \textbf{T + A} & 0.2118 & 0.1841 & 0.1903 \\
& \textbf{T + A + C + IR} & 0.2554 & 0.2360 & 0.2367 \\
 & \textbf{T + A + TI} & 0.2655 & 0.2313 & 0.2407 \\ 
 \bottomrule
\end{tabular}%
}
\caption{Performance of GPT-4o and Claude Sonnet-3.7 on schema generation, given varying paper inputs (T: titles, A: abstracts, TI: table intents, FT: full-texts, C: table captions, IR: in-text references). Incorporating table intents improves F1 AUC consistently across models.} 
\label{tab:intent_results}
\end{table}

\section{Improving Schema Generation: Prompting and Finetuning}
\label{sec:generating_schemas}
With this improved task specification and evaluation strategy, we explore methods for better schema generation. We assess two categories of methods: (i) prompting-based methods that decompose schema generation into atomic sub-tasks (\S\ref{sec:prompting_for_schemas}), and (ii) fine-tuning language models (\S\ref{sec:training_for_schemas}). We evaluate all methods on \datasetname as well as \harddatasetname, a subset we curate as described in \S\ref{ssec:harddataset}, and report results in \S\ref{sec:generation_results}. 


\subsection{Curating \harddatasetname}
\label{ssec:harddataset}
We observe that instances in \datasetname sometimes contain one of the following issues:
\begin{itemize}[leftmargin=*,itemsep=-0.5em]
    \item \emph{Generic} columns (e.g., year of publication, research focus etc.)
    \item \emph{Unrecoverable} columns containing information that cannot be obtained from full-texts of papers in the table (e.g., dataset instances)
\end{itemize}
Generic columns are trivially easy to generate (over-optimistic performance estimates), while unrecoverable columns are impossible to generate (under-optimistic estimates). Therefore, evaluating on a subset free from these issues ensures that we obtain a realistic estimate of model performance. Since filtering such instances automatically is non-trivial, we manually curate \harddatasetname to be free of these issues. We first filter out all tables that have $<5$ papers and $<4$ columns, leaving us with 370 instances. We manually go through these instances, and discard any examples with >=50\% of generic or unrecoverable columns, or leakage of table information into the caption, resulting in 170 instances. We randomly sample 100 of these tables to create \harddatasetname, which will be released on publication. 

\subsection{Prompting for schema generation}
\label{sec:prompting_for_schemas}

\begin{table*}[]
\resizebox{\textwidth}{!}{%
\begin{tabular}{@{}ccclcccccc@{}}
\toprule
\multicolumn{1}{l}{} & \multicolumn{1}{l}{} &  &  & \multicolumn{3}{c}{\textbf{\datasetname}} & \multicolumn{3}{c}{\textbf{\harddatasetname}} \\ \midrule
\textbf{Method} & \textbf{Model} & \textbf{\baselinesname} & \textbf{Input} & \textbf{Recall} & \textbf{Precision} & \textbf{F1} & \textbf{Recall} & \textbf{Precision} & \textbf{F1} \\ \midrule
\multirow{10}{*}{\textbf{Prompting}} & \multirow{5}{*}{\textbf{GPT-4o}} & Joint & \textbf{T + A + TI} & 0.2811 & 0.2648 & 0.2666 & 0.3015 & 0.3011 & 0.2965 \\
 &  & Joint & \textbf{T + FT + TI} & 0.2761 & 0.2518 & 0.2566 & 0.3003 & 0.2982 & 0.2948 \\
 &  & Joint & \textbf{T + A + TI + ICL} & 0.2941 & 0.2807 & \textbf{0.2811} & 0.3126 & {0.3152} & \textbf{0.3096} \\
 &  & Parallel & \textbf{T + FT + TI} & 0.2569 & 0.2407 & 0.2424 & 0.2925 & 0.2898 & 0.2885 \\
 &  & Sequential & \textbf{T + FT + TI} & 0.3122 & 0.2297 & 0.2541 & \textbf{0.3322} & 0.2505 & 0.2765 \\ \cmidrule(l){2-10} 
 & \multirow{5}{*}{\textbf{Sonnet-3.7}} & Joint & \textbf{T + A + TI} & 0.2655 & 0.2313 & 0.2407 & 0.2808 & 0.2542 & 0.2628 \\
 &  & Joint & \textbf{T + FT + TI} & 0.2461 & 0.2299 & 0.2278 & 0.2686 & 0.2550 & 0.2544  \\
 &  & Joint & \textbf{T + A + TI + ICL} & 0.2903 & 0.2505 & 0.2617 & 0.2975 & 0.2742 & 0.2805 \\
 &  & Parallel & \textbf{T + FT + TI} & 0.2659 & 0.2064 & 0.2248 & 0.2775 & 0.2222 & 0.2419 \\
 &  & Sequential & \textbf{T + FT + TI} & \textbf{0.3138} & 0.2090 & 0.2386 & 0.3094 & 0.2075 & 0.2370 \\ \midrule
\multirow{2}{*}{\textbf{Fine-tuning}} & \textbf{Qwen-2.5-3B-Instruct} & Joint & \textbf{T + A + TI} & 0.2638 & \textbf{0.3143} & 0.2772 & 0.2550 & \textbf{0.3221} & 0.2751  \\ \cmidrule(l){2-10} 
 & \textbf{Llama-3.2-3B-Instruct} & Joint & \textbf{T + A + TI} & 0.2684 & 0.3006 & 0.2747 & 0.2653 & 0.3054 & 0.2756 \\ \bottomrule
\end{tabular}%
}
\caption{\small{Schema generation results on \datasetname and \harddatasetname for prompting (\S\ref{sec:prompting_for_schemas}) and fine-tuning methods (\S\ref{sec:training_for_schemas}). Input refers to information from the papers being compared (T: titles, A: abstracts, TI: table intents, FT: full-texts, ICL: in-context learning examples). Task decomposition refers to whether the method sees information from different papers jointly, in parallel, or in sequence. Joint prompting with ICL results in the highest F1 AUC. Prompting baselines achieve higher recall, while fine-tuned models are more precise. Fine-tuned open-weight models are competitive with black-box LLMs. 
}
}
\label{tab:generation_results}
\end{table*}
Schema generation is a complex task that involves processing multiple long input texts. To achieve optimal LLM performance in such cases, researchers often decompose the task into a workflow of several more manageable subtasks \citep{khotdecomposed}. We explore a spectrum of prompting workflows for schema generation, ranging from a straightforward single-step prompt that includes all input papers at once \citep[as in][]{newman2024arxivdigestables}, to multi-step workflows that decompose the task in different ways (sequential or parallel).  

\paragraph{Joint prompting.}
The first class of prompting methods jointly provide the model with information from all $M$ papers to generate a schema in a single step, similar to the baselines in \citet{newman2024arxivdigestables}. 
We first experiment with two variants that provide different kinds of information about input papers, abstracts vs. full-texts. This investigates whether including richer content from the full texts improves performance given the increased context length. Prompts for these setups are provided in \S\ref{sec:prompt_title_abs_intent} and \S\ref{sec:prompt_title_full_text_intent}. We also compare to a variant that includes in-context learning (ICL) examples for the task. From the ``medium quality'' subset of examples released by \citet{newman2024arxivdigestables}, we randomly sample $5$ examples and provide these along with the prompt provided in \S\ref{sec:prompt_icl_title_abs_intent}. 
\paragraph{Parallel prompting.}
In contrast to providing all papers at once, we test a workflow that considers each paper independently, inspired by prior work on concept induction from documents. Specifically, we adapt the Lloom \citep{lam2024concept} system which isolates concepts relevant to the table intent from each paper through a pipeline detailed in \S\ref{sec:appendix_lloom_details}.\footnote{Here, a concept refers to a single sentence or phrase that discusses experimental design or findings from the work.} We aggregate the final concepts from all papers and prompt the model to synthesize them into a single coherent schema of comparison. This decomposes schema generation into simpler subproblems, enabling the use of full-text information without requiring the model to handle entire paper(s) in context, thereby mitigating potential issues related to long-context processing \cite{liu2024lost}. We provide details of the whole pipeline including the prompts for each step in \S\ref{sec:appendix_lloom_details}.\footnote{We also ablate the use of the table intent at various parts of the Lloom pipeline in \S\ref{sec:appendix_lloom_details}, with the highest-scoring variant's scores reported in \S\ref{tab:generation_results}.}

\paragraph{Sequential prompting.}
Finally, we explore a prompting workflow that iteratively processes information from papers in batches as proposed in \cite{wang2025can}. We first prompt the model to summarize the content of the papers from the \emph{titles and full text}. We then start with an empty schema and update it iteratively as each batch is introduced. At each step, we prompt the model with the current schema, the table intent, and a batch of paper summaries to update the schema by adding, removing, or modifying elements based on the new information. This process is repeated for five passes over the set of papers, following the setup in \citet{wang2025can}. The prompts used for this pipeline are provided in \S\ref{sec:appendix_sequential_prompting_details}.

\subsection{Fine-tuning for schema generation}
\label{sec:training_for_schemas}
Besides assessing a wide range of prompting techniques, we also explore the viability of fine-tuning smaller models for this task.  Training smaller local models offers a potentially improved cost/accuracy tradeoff, and can enable schema generation over private or paywalled document collections. For these experiments, we use the ``medium-quality'' subset of 22k tables released by \citet{newman2024arxivdigestables}, referred to as \silverdatasetname. We fine-tune open-weight models from the Llama-3.2 \citep{grattafiori2024llama} and Qwen-2.5 model families \citep{qwen2025qwen25technicalreport} to generate schemas given information from the papers being compared. We train all models to process input papers \emph{jointly}, providing paper titles, abstracts and table intents as input, and to generate the schema in JSON format.
\footnote{We provide additional details about the dataset creation, tokenization, fine-tuning and inference in \S\ref{sec:appendix_training_details}.}

\subsection{Results}
\label{sec:generation_results}
We report all prompting and fine-tuning results for schema generation in \Cref{tab:generation_results}. 

\paragraph{Full paper texts don't move the needle on schema generation.} On both \datasetname and \harddatasetname, all methods using full-texts as input underperform joint prompting with titles, abstracts and table intents on F1 AUC, irrespective of the workflow used. This provides some evidence that connecting information from long context to provide accurate output for complex tasks remains challenging for contemporary LLMs \citep{gao2024insights}.
\paragraph{Sequential prompting promotes high recall, while joint prompting balances high recall and precision.} Both Lloom and sequential prompting underperform joint prompting on F1 AUC. However, sequential prompting achieves the highest recall of all methods on average. This suggests this method tends to favor adding schema elements, leading to broader coverage at the cost of precision.
\vspace{-2.0em}
\paragraph{ICL examples help performance at a high cost.} Joint prompting with in-context examples and user goals with GPT-4o achieves the highest F1 AUC across all methods, highlighting the benefit of providing task demonstrations.\footnote{We confirm that this result holds with significance at the 5\% level of a two-tailed t-test in \S\ref{sec:appendix_significance_generation}.} Among the Claude prompting baselines as well, the highest performance is with ICL examples. However, we note that this comes at a high cost as context lengths for this method are significantly longer than the other joint prompting baselines---the prompt length is $14k$ tokens on average when using ICL examples vs $5.6k$ tokens when not using them.
\vspace{-0.6em}
\paragraph{Fine-tuned smaller, open-weight models can compete with black-box models.}
From \S\ref{tab:generation_results}, we see that fine-tuned models are competitive with prompting-based methods, outperforming the corresponding GPT-4o variant (joint prompting with titles, abstracts, and table intents) on F1 AUC, despite being an order of magnitude smaller in size. While prompting methods obtain higher recall, fine-tuned models achieve higher precision than recall, with Qwen-2.5-3B obtaining the highest precision AUC on both \datasetname and \harddatasetname.  
\section{Refinement of Schemas}
\label{sec:editing_schemas}

\begin{figure}
    \centering
    \includegraphics[width=\linewidth]{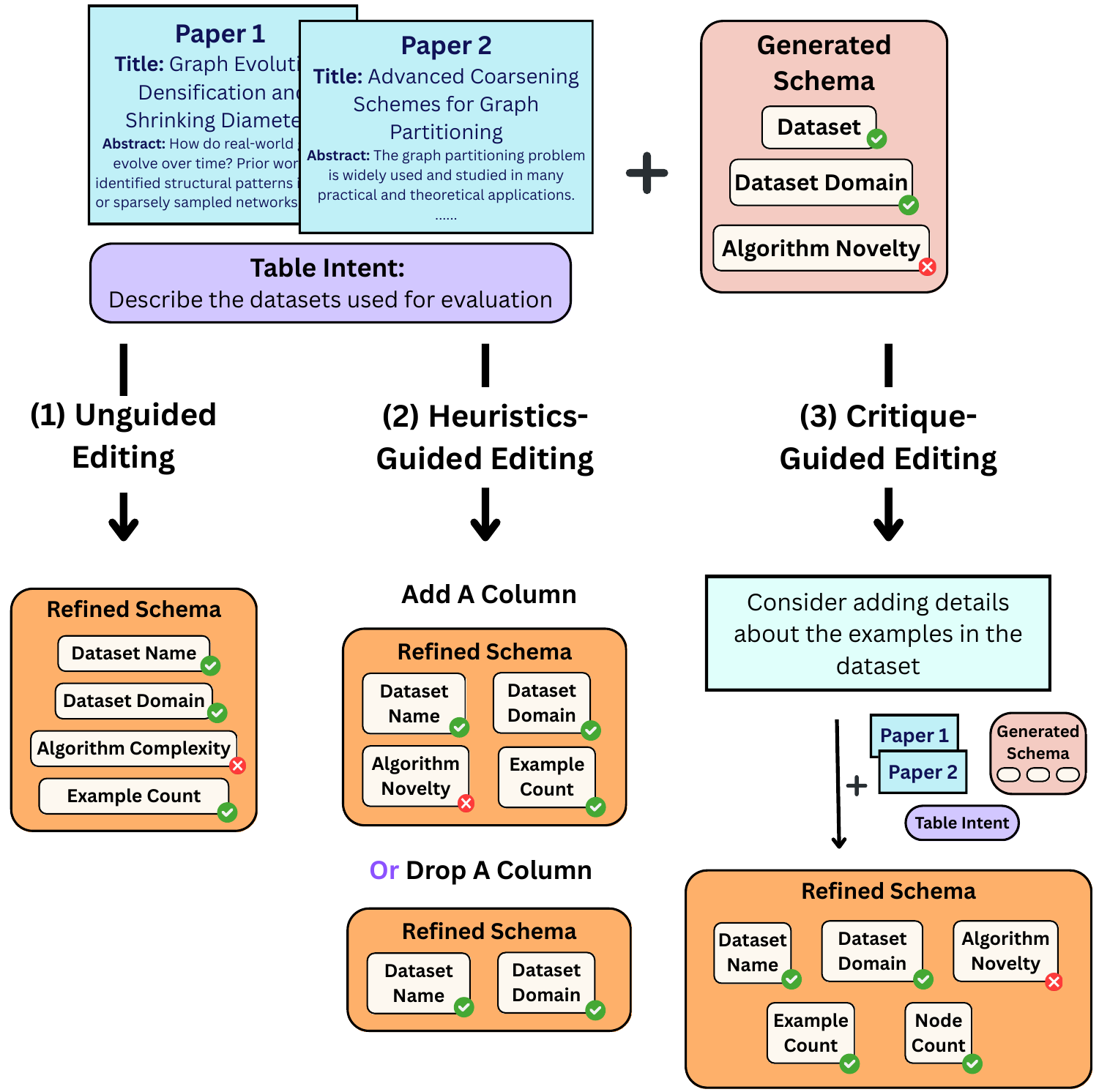}
    \caption{Overview of editing techniques proposed to refine schemas in \S\ref{sec:editing_schemas}. 
    }
    \label{fig:editing_fig1}
\end{figure}

Motivated by the iterative process humans follow while creating schemas \cite{fok2025facetstaxonomiessynthesesnavigating}, which reflects the open-ended nature of the task, we propose LLM-based editing techniques to further refine schemas. These techniques also naturally lend themselves to the development of interactive systems in which users could intervene and guide edits, providing further motivation to explore these techniques. We evaluate three broad categories of refinement strategies, based on how edits are chosen and represented---\emph{unguided} editing without any intermediate description or control (\S\ref{sec:residual_editing}); \emph{heuristics-guided} editing, where edits can only be chosen by a user from fixed set of operations like adding or dropping a column from the schema (\S\ref{sec:heuristics_editing}); and \emph{critique-guided} editing, where free-form natural language critiques specify which edits to make (\S\ref{sec:critiques_editing}). Figure~\ref{fig:editing_fig1} briefly summarizes these editing strategies.

\begin{table*}[]
\resizebox{\textwidth}{!}{%
\begin{tabular}{@{}ccrcccccc@{}}
\toprule
\multirow{2}{*}{\textbf{Method}} & \multirow{2}{*}{\textbf{Model}} & \multirow{2}{*}{\textbf{Input}} & \multicolumn{3}{c}{\textbf{\datasetname}} & \multicolumn{3}{c}{\textbf{\harddatasetname}} \\ \cmidrule(l){4-9} 
 &  &  & \textbf{Recall} & \textbf{Precision} & \textbf{F1} & \textbf{Recall} & \textbf{Precision} & \textbf{F1} \\ \midrule
\textbf{Baseline} & \textbf{GPT-4o} & \textbf{T + A + TI} & 0.2811 & 0.2648 & 0.2666 & 0.3015 & 0.3011 & 0.2965 \\ \midrule
\multirow{2}{*}{\textbf{\begin{tabular}[c]{@{}c@{}}Unguided \\ Editing\end{tabular}}} & \textbf{Llama 3.2 3B Instruct} & \textbf{T + A + TI} & 0.2736* & \textbf{0.2979*} & 0.2764* & 0.2747 & 0.3045 & 0.2803 \\ \cmidrule(l){2-9} 
 & \textbf{Qwen 2.5 3B Instruct} & \textbf{T + A + TI} & 0.2680* & 0.2855* & 0.2643 & 0.2721 & \textbf{0.3252} & 0.2856 \\ \midrule
\multirow{4}{*}{\textbf{\begin{tabular}[c]{@{}c@{}}Heuristics\\ Guided\\ Editing\end{tabular}}} & \multirow{2}{*}{\textbf{Llama 3.2 3B Instruct}} & \textbf{T + A + TI + AC} & 0.2958* & 0.2641 & 0.2723* & 0.3051 & 0.2983 & 0.2976 \\
 &  & \textbf{T + A + TI + DC} & 0.2763* & 0.2765* & 0.2688 & 0.2883 & 0.3065 & 0.2972 \\ \cmidrule(l){2-9} 
 & \multirow{2}{*}{\textbf{Qwen 2.5 3B Instruct}} & \textbf{T + A + TI + AC} & \textbf{0.3031*} & 0.2700 & \textbf{0.2792*} & \textbf{0.3113} & 0.3030 & \textbf{0.3021} \\
 &  & \textbf{T + A + TI + DC} & 0.2844 & 0.2769* & 0.2736* & 0.2930 & 0.3071 & 0.2940 \\ \midrule
\multirow{5}{*}{\textbf{\begin{tabular}[c]{@{}c@{}}Critique\\ Guided\\ Editing\end{tabular}}} & \multirow{3}{*}{\textbf{GPT-4o}} & \textit{\textbf{T + A + TI + RS (Oracle)}} & \textit{0.3393*} & \textit{0.2916*} & \textit{0.3067*} & \textit{0.3438} & \textit{0.3201} & \textit{0.3274} \\
 &  & \textbf{T + A + TI (Self-Refine)} & 0.2836 & 0.2558* & 0.2627 & 0.2995 & 0.2880 & 0.2890 \\
 &  & \textbf{T + A + TI + ICL (Self-Refine)} & 0.2788 & 0.2436* & 0.2533* & 0.2934 & 0.2810 & 0.2829 \\ \cmidrule(l){2-9} 
 & \textbf{Llama 3.2 3B Instruct} & \textbf{T + A + TI (Distilled)} & 0.2917* & 0.2594* & 0.2682 & 0.3041 & 0.2878 & 0.2991 \\ \cmidrule(l){2-9} 
 & \textbf{Qwen 2.5 3B Instruct} & \textbf{T + A + TI (Distilled)} & 0.2916* & 0.2608 & 0.2690 & 0.2953 & 0.2876 & 0.2857 \\ \bottomrule
\end{tabular}%
}
\caption{\small{Evaluating different editing methods for refining candidate schemas (\S\ref{sec:editing_schemas}) on \datasetname and \harddatasetname (Recall, Precision and F1 AUC). We edit the T+A+TI joint prompting baseline with GPT-4o from \S\ref{tab:editing_results}, each subsequent row edits schemas from it using paper information (T:paper titles, A: abstracts, TI: table intents, ICL: in-context learning examples, AC: \emph{Add Column}, DC: \emph{Drop Column}). Cells marked with an asterisk differ from the baseline row with statistical significance at the 5\% level of a two-tailed t-test. Unguided editing shows limited improvements, heuristics consistently improve schemas and critique-guided editing, while helpful in the oracle setting, does not perform well without access to reference schemas.
}
}
\label{tab:editing_results}
\end{table*}

\subsection{Unguided editing}
\label{sec:residual_editing}
We first evaluate whether models can refine schema candidates without generating explicit intermediate feedback or instruction. The model receives as input a candidate schema and information from the papers being compared, and is trained to directly generate a refinement (see Figure~\ref{fig:editing_fig1}). To construct fine-tuning data for this, we use the outputs from joint prompting with GPT-4o with the T+A+TI setup (\Cref{sec:prompting_for_schemas} and \Cref{tab:generation_results}) as candidate schemas on \silverdatasetname examples. These are paired with their corresponding table intents, paper titles and abstracts, and the target output is the reference schema. We fine-tune Llama 3.2-3B-Instruct and Qwen-2.5-3B-Instruct 
to generate the reference schema tokens, conditioned on this input.\footnote{We provide additional fine-tuning details in \S\ref{sec:appendix_unguided_editing}.} 

\subsection{Heuristics-guided editing}
\label{sec:heuristics_editing}
Purely data-driven editing with no intermediate feedback is limited in that it offers no control over the editing process. To address this, we explore whether models can be trained to perform fixed categories of edit operations reliably, which offers the option to introduce user intervention in the refinement process. In our experiments, we focus on two basic operations---adding a column to a schema (\emph{AC}) and dropping a column from a schema (\emph{DC}). 

To train models for \emph{AC}, we create training instances by randomly dropping one column from reference schemas in \silverdatasetname. The input includes the incomplete schema, table intent, titles, and abstracts; the target is the original schema (Figure~\ref{fig:editing_fig1}). For \emph{DC}, we append a column from a GPT-4o generation to the reference schema, creating input with a potentially noisy extra item. The model sees the full context and learns to recover the original schema.
We fine-tune Llama 3.2-3B-Instruct and Qwen 3.2-3B-Instruct on datasets for both operations to serve as heuristic-guided editors and report the results in \S\ref{tab:editing_results}.\footnote{We provide additional fine-tuning details in \S\ref{sec:heuristics_editing}.}

\subsection{Critique-guided editing}
\label{sec:critiques_editing}
Lastly, we evaluate whether LLMs can refine schema candidates based on generated natural language critiques. A critique takes the form of a sentence describing an issue with the current schema and how it should be edited to better align with the table intent and input papers (e.g., ``Add fine-grained detail about datasets such as number of examples''). This 
allows for more expressive edits than the atomic operations explored in \S\ref{sec:heuristics_editing}. 
We explore three types of critique-guided approaches (summarized by Figure~\ref{fig:critiques} in \S\ref{sec:appendix_prompting_critiques_details}):
\begin{itemize}[leftmargin=*,itemsep=-0.5em]
    \item \textbf{Oracle:} To test whether models can generate and implement critiques effectively, we begin with an \emph{oracle} setting. Given a candidate schema, reference schema, and information from the compared papers, we prompt GPT-4o to generate a single natural language critique, then update the schema using this critique. Since critiques are generated with access to the reference, they should lead to edits that improve performance. 
    \item \textbf{Self-Refine:} We evaluate a \emph{self-refinement} setting inspired by \citet{madaan2023self}. Starting with a candidate schema and information from compared papers, we prompt GPT-4o to produce a single, atomic critique, without access to the reference; then prompt it again to update the schema according to the critique. Prompts used for each step are provided in (\S\ref{sec:appendix_prompting_critiques_details}). We also evaluate a \emph{self-refinement setting with in-context examples} of critiques. In-context critiques are chosen by retrieving the $5$ most similar examples from \silverdatasetname based on BERTScore similarity between their table intents \citep{Zhang2020BERTScore} and adding oracle critiques for these examples to the prompt provided in \S\ref{sec:appendix_prompting_critiques_details}.
    \item \textbf{Distilled:} We investigate whether critique generation skills can be distilled into open-weight models. We construct fine-tuning data using oracle critiques for \silverdatasetname, where the input is a candidate schema and paper input (\emph{T+ A + TI}), and the output is the corresponding oracle critique. 
    We fine-tune Llama 3.2-3B-Instruct and Qwen 2.5-3B-Instruct for critique generation using supervised fine-tuning, and prompt GPT-4o to revise schemas using critiques generated by the distilled models.\footnote{\S\ref{sec:appendix_prompting_critiques_details} provides additional fine-tuning details.}
\end{itemize}
\subsection{Results}
\label{sec:editing_results}
\paragraph{Unguided editing offers limited improvement.} \S\ref{tab:editing_results} shows that Llama 3.2-3B-Instruct trained for unguided editing improves candidate schemas leading to better F1 AUC on \datasetname, but Qwen 2.5-3B-Instruct shows no improvement. Additionally, both models show performance drops on \harddatasetname, indicating that unguided editing might be difficult for models to learn. 
\paragraph{Heuristics-guided edits improve performance as directed.}
From \S\ref{tab:editing_results}, we find that \emph{Add} 
(AC) operations increase recall while slightly lowering precision, as expected when an additional schema item is introduced. Conversely, \emph{Drop} (DC) operations raise precision at the cost of recall, showing that the model effectively removes less relevant elements. This highlights the potential of incorporating controllable atomic edits in an interactive setup, allowing users to specify different schema qualities to prioritize, depending on their goals. 
\paragraph{Oracle critiques consistently improve schemas, but Self-Refine proves challenging.} \S\ref{tab:editing_results} shows that critique-guided editing in the oracle setting improves performance on all metrics, indicating that LLMs can leverage good natural language feedback to produce more accurate schemas. Self-refinement, however, leads to a slight drop in performance compared to the original schemas, even when ICL examples are used. A manual qualitative error analysis (\S\ref{sec:critiques_error_analysis}) over 25 examples of generated critiques shows that this is largely due to critiques that are overly generic (24\%) or contain small factual errors (20\%) or mistakes in reasoning (20\%), which lead to sub-optimal edits. 
\paragraph{Distilling critique-generation works better with room to improve.}
As shown in \S\ref{tab:editing_results}, critiques from distilled models improve recall and yield modest F1 AUC gains on \datasetname and \harddatasetname. This highlights a promising future research direction on using LLMs for critiquing challenging reasoning outputs. 

\section{Related Work}
\label{sec:related}

\paragraph{Schema Generation as a Task} The task of creating schemas to compare multiple documents has been studied extensively \citep{shahaf2012metro, zhang2018fly, zhu-etal-2023-hierarchical}. Several efforts have also focused on schema generation specifically for research papers, using various structures (e.g., tables, outlines, etc.) to represent schemas \citep{Hashimoto2017AutomaticGO, gupta2023discomat, bai2024schema, zhu-etal-2023-hierarchical, hsu2024chime, newman2024arxivdigestables}. Given recent progress on generating tables from text \citep{parikh2020totto, wu2022text, tang2023struc, sundar2024gtbls}, we adopt the table-based schema representation proposed by \citet{newman2024arxivdigestables}.

A common limitation in recent datasets is the absence of an explicit input query or user intent, identified as important by earlier work \citep{zhang2018fly}, which directs schema generation and helps disambiguate between multiple valid candidate schemas. In this work, we augment the dataset from \citet{newman2024arxivdigestables} with synthetic table intents that serve this disambiguating function. Concurrent work from \citet{wang2025can} independently make a similar observation and define \emph{user demands} to guide schema generation. Our work differs in that we also benchmark a range of schema creation and editing methods—including the prompting strategy proposed by \citet{wang2025can}—whereas their focus is primarily on improving task definition and evaluation by adding user demands and distractor papers to the task. 

\paragraph{Schema Generation Methods}  The schema generation methods we benchmark are motivated by findings that in-context learning can improve performance on complex tasks \citep{brown2020language}, as well as by recent work that leverages multi-step prompting workflows to support better concept induction \citep{lam2024concept} and schema construction \citep{wang2025can}. 
Our work is the first to evaluate methods to edit schema generation candidates. We note that our experiments with critique-based revisions are informed by self-refinement literature \citep{madaan2023self}, as well as evidence of strong performance from skill distillation into open-weight models 
\citep{hinton2015distilling, alpaca, wang2023self}. 
\section{Conclusion}
\label{sec:conclusion}
Large language models have the potential to serve as valuable tools for literature review by automatically synthesizing schemas for comparing research papers. In this work, we show that existing formulations of the task miss an important component, an explicit specification of table intent. We show that augmenting \datasetname with synthetically generated table intents significantly improves schema generation performance when prompting LLMs. We also demonstrate that fine-tuning open-weight models that match the performance of state-of-the-art black-box LLMs at a fraction of the model size. We also explore methods for refining schema candidates---unguided, heuristic-guided, and critique-guided edits---and show that open-weight models can be fine-tuned to further improve schemas. We note limitations in the abilities of LLMs to critique existing candidates, highlighting an important direction of future work. To support future work, we release all fine-tuned models, our augmented \datasetname, and the manually curated \harddatasetname.

\section*{Acknowledgements}
We would like to thank Benjamin Newman, Nishant Balepur, Nitish Joshi, Yoonjoo Lee, Nick Lourie, and members of the Semantic Scholar research group for their feedback at various stages of the project. This work was completed while Vishakh was an intern at AI2. At NYU, Vishakh is supported by the National Science Foundation under Grant No. IIS-2340345 and Grant No. 1922658. 

\section*{Limitations}
\label{sec:limitations}
Our experiments are conducted primarily on research papers from arXiv. While the topics covered are broad, this focus limits the generalizability of our findings to domains that commonly publish on the platform. Additional work is needed to assess whether the results extend to other types of documents.

We evaluate schema generation using BERTScore to align predicted schema items with references, following the setup in \citet{newman2024arxivdigestables}. However, string similarity metrics like BERTScore have known limitations \citep{sun2022bertscore, chen2022reproducibility}, and may not fully capture semantic alignment. While more reliable, human evaluations are costly as they would require expert annotators to be recruited for extended periods of time making them infeasible for large-scale experiments. We recommend incorporating human studies in downstream or user-facing deployments.

Our experiments use a JSON-based schema format for its readability and ease of parsing, though this may not be the optimal representation. Lastly, we use black-box LLMs (GPT-4o, Claude Sonnet) for analysis. Although these models are generally reliable, their proprietary nature introduces concerns about reproducibility. To address this, we also explore fine-tuning open-weight models.


\bibliography{custom}

\appendix
\section{Open Source Research Materials}
\label{sec:released_materials}
\subsection{Data}

We release an updated version of \texttt{\datasetname} along with table intents as a HuggingFace Dataset - \url{https://huggingface.co/datasets/vishakhpk/ArxivDIGESTables-v0.1}.

We also release the filtered, manually curated subset \texttt{\harddatasetname} on HuggingFace - \url{https://huggingface.co/datasets/vishakhpk/ArxivDIGESTables-Clean}.

\subsection{Models}

We also the fine-tuned models listed in \Cref{tab:model_directory} on HuggingFace.

\begin{table*}[h]
\centering
\begin{tabular}{cC{9cm}C{3cm}}
\toprule
\textbf{Task} & \textbf{Model Name} & \textbf{Model on HF} \\
\midrule
Schema Generation & Joint Generation - T+A+TI & 
\href{https://huggingface.co/vishakhpk/schema_generation_llama_3.2_3b_instruct}{Llama} and 
\href{https://huggingface.co/vishakhpk/schema_generation_qwen_2.5_3b_instruct}{Qwen} \\

Schema Editing & Unguided Editing - T+A+TI & 
\href{https://huggingface.co/vishakhpk/schema_editing_residual_llama_3.2_3b_instruct}{Llama} and 
\href{https://huggingface.co/vishakhpk/schema_editing_residual_qwen_2.5_3b_instruct}{Qwen} \\

Schema Editing & Drop Column - Heuristics-Guided Editing - T+A+TI & 
\href{https://huggingface.co/vishakhpk/schema_editing_drop_column_llama_3.2_3b_instruct}{Llama} and 
\href{https://huggingface.co/vishakhpk/schema_editing_drop_column_qwen_2.5_3b_instruct}{Qwen} \\

Schema Editing & Add Column - Heuristics-Guided Editing - T+A+TI & 
\href{https://huggingface.co/vishakhpk/schema_editing_add_column_llama_3.2_3b_instruct}{Llama} and 
\href{https://huggingface.co/vishakhpk/schema_editing_add_column_qwen_2.5_3b_instruct}{Qwen} \\

Schema Editing & Critique Generation - T + A + TI (Distilled) & 
\href{https://huggingface.co/vishakhpk/schema_editing_critique_generation_llama_3.2_3b_instruct}{Llama} and 
\href{https://huggingface.co/vishakhpk/schema_editing_critique_generation_qwen_2.5_3b_instruct}{Qwen} \\
\bottomrule
\end{tabular}
\caption{Finetuned Schema Generation and Editing LLMs with HuggingFace links.}
\label{tab:model_directory}
\end{table*}

\section{Creating table intents}
\label{sec:appendix_user_goals}
The model we use to create table intents using \texttt{gpt-4o-2024-08-06} accessed between August 2024 and March 2025. We first prompt the model with the table, caption and in text references to create table intents using the prompt in \S\ref{sec:prompt_create_user_goals}. We sample $5$ candidates at temperature $0.7$. We then score the candidates and select the best one using LLM-as-judge with the prompt in \S\ref{sec:prompt_select_user_goals} which accepts the information about the table as well as all the candidate intents concatenated together. To confirm the validity of the LLM-as-judge step, we perform a human annotation on a random subset of $30$ examples. We provide annotators with the set of $5$ candidate intents as well as the specification from \S\ref{sec:prompt_create_user_goals}. We ask these annotators to rank the set of intents, collecting $3$ different annotations per example. We then compute the Spearman rank correlation between the LLM-as-judge ratings and the annotations. We observe an average correlation of $0.73$ with standard deviation $0.08$ which corresponds to a medium to high correlation between the rankings, confirming the validity of our created intents. 

\subsection{Prompt to create user goals}
\label{sec:prompt_create_user_goals}

\tiny
\begin{verbatim}
GENERATE_SYNTHETIC_GOALS_PAPERS_QUESTION = '''
When writing a scientific research paper, we often include
tables comparing different works to accomplish a variety
of goals. 
The author has this goal in mind when they create the table
for what they want to convey to the reader via the
objective comparison of papers. \
For example, some potential goals might include: \
1. Highlighting gaps in existing research: By comparing
related studies, the table can show areas where there is
limited research or unresolved questions, positioning the
current study as addressing those gaps. \
2. Contextualizing the study: It helps place the current
research within the broader scientific context, showing
how it builds upon or differs from previous work. \
3. Evaluating methodology differences: It allows for an
easy comparison of the methodologies used in different
studies, illustrating why the chosen methods in the
current paper are innovative, more robust, or better 
suited for the research problem. \
4. Demonstrating novelty: By showing what has already
been done, a comparison table emphasizes the unique
contribution or novelty of the present study. \
5. Assessing the consistency of results: The table can
highlight differences or consistencies in findings across
studies, helping the reader understand how results align
or contrast with existing literature. \
6. Simplifying complex information: It makes it easier for 
readers to quickly grasp how various studies relate to one
another, especially when reviewing large bodies 
of literature. \
7. Supporting the literature review: It strengthens the
literature review by systematically summarizing relevant
research, which aids in the argument for why the current 
study is needed. \
Generally this goal can be written down as a simple 
open-ended question that the author anticipates that 
the reader will have and that can be answered with the table. \
Your task is to generate this goal given a particular table
from a research paper. You are also given the title 
and abstract of the paper, the description of the table 
and additional information about how the table is 
referenced in the text of the paper. \
[Table] {table}
[Caption] {caption} \
[In-text references] {in_text_refs} \
We also provide information about the papers being discussed
in the table. You want the goal to be one that helps a 
future user actionably create the table given the information
in these papers: \
{papers}
Return output in the following JSON format: 
{{'goal':<your goal>, 'justification':<justification of the
goal>}}
'''
\end{verbatim}

\subsection{Prompt to pick the best user goal}
\label{sec:prompt_select_user_goals}

\tiny
\begin{verbatim}
EVALUATE_GOALS_TO_TABLE = '''
Imagine you are a co-author of a scientific paper and the
first author has created a table comparing different 
papers/methods. You are reading the table along with the
caption of the paper and references to the table in the 
text of the paper. You are trying to guess what is the 
intent with which your co-author created this particular 
table. \
Given a set of candidate intents that you think they 
might have had, your task is to select the best user 
intent out of them. Assign a score to each candidate on
a scale of 1 to 5 on how well it fits what they might have
thought. Prioritize selecting a user intent that is highly
specific to the particular information in the table. 
The output format is a JSON with a string valued 
justification containing the scores assigned to each 
candidate schema along with why that score was assigned. 
You should also provide your final choice of the best 
schema. If you feel that none of them are good, then 
reply with None here. \
[Table] {table} \
[Caption] {caption} \
[In-text references] {in_text_refs} \
[Candidate goals] {goal_text} \
Return the output in the following JSON format. The
justification should include the reasoning for the 
score as a string, the best_goal should be the text of 
the best candidate and nothing else: {{'justification':
<justification for the score>, 'best_goal':<the best 
candidate selected>}} \
'''
\end{verbatim}

\section{Example of a schema as JSON}
\label{sec:appendix_example_schema_json}
\normalsize
Consider the example table in \S\ref{tab:example_schema_for_json}. The JSON version of this schema is as follows:
\tiny
\begin{verbatim}
{
    "GAN": {
        "definition": "The name of the 
        Generative Adversarial Network model 
        being evaluated.",
        "output_format": "string values"
    },
    "Dataset": {
        "definition": "The name of the dataset
        used for training and evaluating 
        the GAN model.",
        "output_format": "string values"
    },
    "# Examples": {
        "definition": "The total number of 
        examples in the dataset used for 
        training and validation of the model.",
        "output_format": "string values 
        (may include numbers formatted 
        with commas)"
    },
    "Classes": {
        "definition": "The number of distinct
        classes present in the dataset.",
        "output_format": "integer values"
    },
    "Class Balance": {
        "definition": "The distribution of 
        examples among the classes in the 
        dataset, expressed as a percentage 
        split.",
        "output_format": "string values, 
        usually in percentage format"
    },
    "Accuracy": {
        "definition": "The classification 
        accuracy achieved by the model on 
        a holdout set from the dataset.",
        "output_format": "percentage 
        values as strings"
    },
    "Source": {
        "definition": "A citation or 
        reference to the source of the model
        or data.",
        "output_format": "string values, 
        formatted as citation keys"
    }
}
\end{verbatim}

\begin{table}[]
\resizebox{\columnwidth}{!}{%
\begin{tabular}{@{}lllllll@{}}
\toprule 
\textbf{GAN} & \textbf{Dataset} & \textbf{\# Examples} & \textbf{Classes} & \textbf{Class Balance} & \textbf{Accuracy} & \textbf{Source} \\ \midrule
BicycleGAN & Edges2Shoes & 300 & 2 & 45\%/55\% & 94\% & \{\{cite:51b5159\}\} \\
AttGAN & CelebA & 900 & 2 & 49\%/51\% & 98\% & \{\{cite:dbd4855\}\} \\
BigGAN & ImageNet & 1281167 (train) & 1000 & Varying & 75\% & \{\{cite:34ba2e5\}\} \\
ShapeHDGAN & ShapeNet & 600 & 2 & 49\%/51\% & 96\% & \{\{cite:ebfd324\}\} \\
StyleGAN2 & Style & 540 & 2 & 49\%/51\% & 98\% & \{\{cite:5aac1d6\}\} \\
cGAN & MNIST & 9000 & 10 & 10\% each & 96\% & \{\{cite:49faa9e\}\} \\
\bottomrule
\end{tabular}%
}
\caption{Example table used to describe the JSON format of our schemas in \S\ref{sec:appendix_example_schema_json}.}
\label{tab:example_schema_for_json}
\end{table}

\section{Prompting Details}
\label{sec:appendix_prompting_hyperparams}
\normalsize
For all prompting experiments, we set the system prompt to the one provided in \S\ref{sec:prompt_system}. Unless stated otherwise, we sample one output per example with temperature $0.7$. For all GPT-4o results, we use \texttt{gpt-4o-2024-08-06} via the API accessed between August 2024 and May 2025. For Claude 3.7 Sonnet, we access the model via the API between February and May 2025. We access DeepSeek R1 via the Together API between February and May 2025. Each prompt is provided in the following sections and referred to in the corresponding sections from the main text. Inputs in these prompts are enclosed within \{ and \} and programmatically inserted from each example. 

\section{Prompts used in \S\ref{sec:generating_schemas}}
\label{sec:joint_prompting_setup}


\subsection{System Prompt for all experiments}
\label{sec:prompt_system}
\tiny
\begin{verbatim}
SYSTEM_PROMPT = "You are an intelligent and precise 
assistant that can understand the contents of research 
papers. You are knowledgable on different fields and
domains of science, in particular computer science. 
You are able to interpret research papers, create 
questions and answers, and compare multiple
papers."
\end{verbatim}

\subsection{Prompt for Title+Abstracts}
\label{sec:prompt_joint_title_abs}
\tiny
\begin{verbatim}
SCITABLES_OPEN_LENGTH_SCHEMA = '''
Imagine the following scenario: A user is making a table
for a scholarly paper that contains information about 
multiple papers and compares these papers. To compare 
and contrast the papers , the user provides the title 
and content of each paper. \
Your task is the following: Given a list of papers , you
should find the appropriate number of attributes that are
shared by the given research papers and can be used to 
compare them. So each attribute is a topic that would be
covered in the Related Work section of the user's paper. \
Return a JSON object in the following format: \"""json {{"
<attribute 1>": {{"definition":<your definition of why this
attribute should be an axis of comparison>, "output_format":
<describe the range of output values that will be filled in,
is it numbers, string values or another format>}} , ...}} 
""" \
{papers} \
'''

\end{verbatim}

\subsection{Prompt for Title + Caption + In-Text Refs}
\label{sec:prompt_cap_intext}
\tiny
\begin{verbatim}
CAPTION_INTEXT_REFS_OPEN_LENGTH_SCHEMA = '''
Imagine the following scenario: A user is making a table
for a scholarly paper that contains information about 
multiple papers and compares these papers. To compare and
contrast the papers , the user provides the title and 
content of each paper. To help you build the table, the 
user provides a caption of this table , which is referred
to in the paper as additional information. \
[Caption] {caption} \
[In-text references] {in_text_refs} \
[Papers] {papers} \
Your task is the following: Given a list of papers, you 
should find the appropriate number of attributes that are 
shared by the given research papers and can be used to 
compare them. So each attribute is a topic that would be 
covered in the Related Work section of the user's paper. \
Return a JSON object in the following format: \"""json 
{{"<attribute 1>": {{"definition":<your definition of why 
this attribute should be an axis of comparison>, 
"output_format":<describe the range of output values that 
will be filled in, is it numbers, string values or 
another format>}} , ...}} """ \
'''
\end{verbatim}

\subsection{Prompt for Title + Abstract + Table Intents}
\label{sec:prompt_title_abs_intent}
\tiny
\begin{verbatim}
GOALS_OPEN_LENGTH_SCHEMA = '''
Imagine the following scenario: A user is making a table
for a scholarly paper that contains information about 
multiple papers and compares them. To compare and contrast
these papers , the user provides the title and content 
of each paper below. To help you build the table , the user
also provides you with the goal that they want to 
accomplish with this table in the form of an open question. \
[User Goal] {user_goal} \
Your task is the following: Given a list of papers , you
should find the appropriate number of attributes that are
shared by the given research papers and can be used to compare
them. So each attribute is a topic that would be covered 
in the Related Work section of the user's paper. Remember,
the table should answer the question from the user goal. \
Return a JSON object in the following format: \"""json 
{{"<attribute 1>": {{"definition":<your definition of why 
this attribute should be an axis of comparison>, 
"output_format":<describe the range of output values that
will be filled in, is it numbers, string values or another
format>}} , ...}} """ \
[Papers] {papers}
'''
\end{verbatim}

\subsection{Prompt for Title+Full Text}
\label{sec:prompt_title_full_text}
\tiny
\begin{verbatim}
FULL_TEXT_OPEN_LENGTH_SCHEMA = '''
Imagine the following scenario: A user is making a
table for a scholarly paper that contains 
information about multiple papers and compares these
papers. To compare and contrast the papers , the 
user provides the title and content of each paper. \
Your task is the following: Given a list of papers , 
you should find the appropriate number of attributes 
that are shared by the given research papers and can
be used to compare them. So each attribute is a topic
that would be covered in the Related Work section of
the user's paper. \
Return a JSON object in the following format: \"""json 
{{"<attribute 1>": {{"definition":<your definition of 
why this attribute should be an axis of comparison>,
"output_format":<describe the range of output values 
that will be filled in, is it numbers, string values 
or another format>}} , ...}} """ \
{full_text_papers} \
'''
\end{verbatim}

\subsection{Prompt for Title+Full Text + Intent}
\label{sec:prompt_title_full_text_intent}
\tiny
\begin{verbatim}
GOALS_FULL_TEXT_OPEN_LENGTH_SCHEMA = '''
Imagine the following scenario: A user is making a 
table for a scholarly paper that contains 
information about multiple papers and compares 
them. To compare and contrast these papers , the
user provides the title and content of each paper
below. To help you build the table , the user also
provides you with the goal that they want to 
accomplish with this table in the form of an open
question. \
[User Goal] {user_goal} \
Your task is the following: Given a list of papers ,
you should find the appropriate number of attributes
that are shared by the given research papers and
can be used to compare them. So each attribute is a
topic that would be covered in the Related Work 
section of the user's paper. Remember, the table 
should answer the question from the user goal. \
Return a JSON object in the following format: \"""json 
{{"<attribute 1>": {{"definition":<your definition of 
why this attribute should be an axis of comparison>,
"output_format":<describe the range of output values 
that will be filled in, is it numbers, string values 
or another format>}} , ...}} """ \
{full_text_papers} \
'''
\end{verbatim}

\subsection{Prompt for ICL examples + Title + Abstract + Intnet}
\label{sec:prompt_icl_title_abs_intent}
\tiny
\begin{verbatim}
ICL_GOALS_OPEN_LENGTH_SCHEMA = '''
Imagine the following scenario: A user is making a table for
a scholarly paper that contains information about multiple
papers and compares them. To compare and contrast these 
papers , the user provides the title and content of each
paper below. Here are some representative examples of the
schema of the table given the content of the papers 
being compared: \
[Representative Examples] {icl_text} \
Your task is the following: Given a list of papers, you 
should find the appropriate number of attributes that are
shared by the given research papers and can be used to 
compare them. So each attribute is a topic that would be
covered in the Related Work section of the user's paper.
Remember, the table should answer the question from the 
user goal. \
Return a JSON object in the following format: \"""json 
{{"<attribute 1>": {{"definition":<your definition of why
this attribute should be an axis of comparison>, 
"output_format":<describe the range of output values that
will be filled in, is it numbers, string values or another
format>}} , ...}} """ \
To help you build the table , the user also provides you
with the goal that they want to accomplish with this table
in the form of an open question. \
[User Goal] {user_goal} \
Here is the information about the papers being compared: \
{papers}
'''
\end{verbatim}

\section{Lloom Details}
\label{sec:appendix_lloom_details}
\normalsize
\subsection{Pipeline of execution}
We adapt Lloom \cite{lam2024concept} for schema generation as follows:
\begin{itemize}
    \item The first step is to summarize the content of the full text of each individual paper being compared into bullet points with the summarize prompts provided in \S\ref{sec:lloom_summarize_prompts}. We experiment with both the original prompt provided in \citet{lam2024concept} as well as a intent-oriented modified version which affects performance as shown in \S\ref{sec:lloom_variant_results}. 
    \item The bullet points obtained from all papers are then converted to embeddings using the \texttt{text-embedding-ada-002} model. This is followed by clustering using HDBSCAN with Euclidean distance as the metric. We set the hyperparameters of this step from \citet{lam2024concept}.
    \item Each cluster is then converted to a concept list using the prompt from \S\ref{sec:lloom_synthesize_prompt}.
    \item The obtained concepts are first filtered for generic or too specific items using the prompt in \S\ref{sec:lloom_filter_prompt} followed by a prompting step that merges similar concepts. We also include a filtering step that removes concepts not relevant to the table intent (prompts provided in \S\ref{sec:lloom_merge_prompt}), ablating this in \S\ref{sec:lloom_variant_results}.
    \item Finally, we convert the filtered concepts into a schema for comparison using the prompt from \S\ref{sec:lloom_concepts_to_schema_prompt}.
\end{itemize}
We ablate the contributions of where to introduce the table intent into the Lloom pipeline in \S\ref{sec:lloom_variant_results}

\subsection{Variants of Lloom}
\label{sec:lloom_variant_results}
We compare three versions of the Lloom pipeline that differ based on how the table intents are incorporated:
\begin{itemize}
    \item Lloom-Lite, which uses the table intent only in the final step for converting the filtered concepts into a. 
    \item Lloom-Lite with all the concepts obtained from the filtering step i.e. we simply create a schema with every concept from here as opposed to prompting an LLM to create a coherent schema from these concepts
    \item Lloom-Lite that also uses the table intent to filter concepts before converting them into the schema. 
    \item Lloom-Lite with filtering of concepts using the table intent, but no other filtering or deduplication.
    \item Lloom-Lite that includes the table intent in the first summarization step in addition to the concept filtering step
\end{itemize}

\paragraph{Results} 
From \S\ref{tab:lloom_variant_results}, we see that incorporating the table intent for concept filtering obtains the highest F1 AUC score---we report this variant in \S\ref{tab:generation_results} in \S\ref{sec:generating_schemas}. Predictably, the all-concept baselines have higher recall, but very low precision as these have a very high number of predicted schema elements. Finally, we see that incorporating the table intent earlier in the Lloom pipeline, directly into the summarization step, leads to lower recall with slightly higher precision.

\begin{table}[]
\resizebox{\columnwidth}{!}{%
\begin{tabular}{@{}rccc@{}}
\toprule
 & Recall & Precision & F1 \\ \midrule
Lloom Lite & 0.2552 & 0.2441 & 0.2417 \\
Lloom Lite - All Concepts & 0.2629 & 0.1206 & 0.1481 \\
Lloom Lite - Intent-Oriented Concept Filtering & 0.2569 & 0.2407 & 0.2424 \\
Lloom Lite - Intent-Oriented Concept Filtering - All Concepts & 0.2631 & 0.1226 & 0.1506 \\
Lloom Lite - Intent Oriented Concept Filtering and Summarization & 0.2363 & 0.2563 & 0.2399 \\ \bottomrule
\end{tabular}%
}
\caption{Variants of Lloom with GPT-4o detailed in \S\ref{sec:lloom_variant_results}}
\label{tab:lloom_variant_results}
\end{table}

\subsection{Lloom prompts}
\tiny
\subsubsection{Summarize prompts}
\label{sec:lloom_summarize_prompts}
\begin{verbatim}
summarize_prompt = """
I have the following TEXT EXAMPLE:
{ex}

Please summarize ALL the text in this EXAMPLE into 
bullet points ensuring that you cover all of the 
content in the example. Each bullet point should 
be a single sentence. The example is a research paper
so make sure you cover all aspects in the various
bullet points including all specific details about
the background, objective, method, experimental 
setups, names of datasets, results and takeaways.
Make sure to be very detailed in writing bullet 
points--for example, don't say 'other methods', 
instead specify which are the methods mentioned in
the text. We want as many details as possible such
that all the information in the paper is covered in
at least one bullet point. Phrase each bullet point
such that it is understandable without needing
external context. Please respond ONLY with a valid
JSON in the following format:
{{
    "bullets": [ "<BULLET_1>", "<BULLET_2>", ... ]
}}
"""

goal_oriented_summarize_prompt = """
I have the following TEXT EXAMPLE:
{ex}

I also have an associated USER INTENT:
{goal}

Please summarize ALL the text in this EXAMPLE into 
bullet points in the context of the provided USER 
INTENT. Make sure that you cover all of the content
in the example, relevant to the particular USER 
INTENT. Each bullet point should be a single sentence.
Tailor each bullet point in the context of providing
information that would be informatinve and relevant to
a user with that USER INTENT. The example is a 
research paper so make sure you cover all aspects in
the various bullet points including all specific 
details about the background, objective, method, 
experimental setups, names of datasets, results and
takeaways. Make sure to be very detailed in writing
bullet points--for example, don't say 'other methods',
instead specify which are the methods mentioned in 
the text. We want as many details as possible such 
that all the information in the paper is covered in at
least one bullet point. Phrase each bullet point such 
that it is understandable without needing external 
context. Please respond ONLY with a valid JSON in 
the following format:
{{
    "bullets": [ "<BULLET_1>", "<BULLET_2>", ... ]
}}
\end{verbatim}

\subsubsection{Synthesize clusters into concepts}
\label{sec:lloom_synthesize_prompt}
\tiny
\begin{verbatim}
synthesize_prompt = """
I have this set of bullet points from a set of 
research papers:
{examples}

Please write a summary of {n_concepts} unifying 
patterns for these examples. {seeding_phrase} For 
each high-level pattern, write a 2-4 word NAME for
the pattern and an associated 1-sentence ChatGPT 
PROMPT that could take in a new text example and 
determine whether the relevant pattern applies. Also
include 1-2 example_ids for items that BEST 
exemplify the pattern. Please respond ONLY with 
a valid JSON in the following format:
{{
    "patterns": [ 
        {{"name": "<PATTERN_NAME_1>", "prompt": "
        <PATTERN_PROMPT_1>", "example_ids": ["<EXAMPLE_ID_1>", "
        <EXAMPLE_ID_2>"]}},
        {{"name": "<PATTERN_NAME_2>", "prompt": "
        <PATTERN_PROMPT_2>", "example_ids": ["<EXAMPLE_ID_1>", "
        <EXAMPLE_ID_2>"]}},
    ]
}}
"""
\end{verbatim}

\subsubsection{Filter prompt}
\label{sec:lloom_filter_prompt}
\begin{verbatim}
review_remove_prompt = """
I have this set of themes generated from text examples:
{concepts}

Please identify any themes that should be REMOVED 
because they are either:
(1) Too specific/narrow and would only describe a few 
examples, or 
(2) Too generic/broad and would describe nearly all 
examples.
If there no such themes, please leave the list empty.
Please respond ONLY with a valid JSON in the 
following format:

{{
    "remove": [ 
        "<THEME_NAME_5>",
        "<THEME_NAME_6>",
    ]
}}
"""
\end{verbatim}

\subsubsection{Merge prompt}
\label{sec:lloom_merge_prompt}

\begin{verbatim}
review_remove_prompt_seed = """
I have this dict of CONCEPTS (keys) and their 
corresponding inclusion criteria (values), as follows:
{concepts}

I have the following THEME:
{seed}

Please identify any CONCEPTS that DO NOT relate to 
the THEME and that should be removed. If there no 
such concepts, please leave the list empty.
Please respond ONLY with a valid JSON in the following format:

{{
    "remove": [ 
        "<CONCEPT_NAME_5>",
        "<CONCEPT_NAME_6>",
    ]
}}
"""

review_merge_prompt = """
I have this set of themes generated from text examples:
{concepts}

Please identify any PAIRS of themes that are 
similar or overlapping that should be MERGED together. 
Please respond ONLY with a valid JSON in the following 
format with the original themes and a new name and 
prompt for the merged theme. Do NOT simply combine the 
prior theme names or prompts, but come up with a new 
2-3 word name and 1-sentence ChatGPT prompt. If there
no similar themes, please leave the list empty.

{{
    "merge": [ 
        {{
            "original_themes": ["<THEME_NAME_A>", "
            <THEME_NAME_B>"],
            "merged_theme_name": "<THEME_NAME_AB>",
            "merged_theme_prompt": "<THEME_PROMPT_AB>",
        }},
        {{
            "original_themes": ["<THEME_NAME_C>", "
            <THEME_NAME_D>"],
            "merged_theme_name": "<THEME_NAME_CD>",
            "merged_theme_prompt": "<THEME_PROMPT_CD>",
        }}
    ]
}}
"""
\end{verbatim}

\subsubsection{Converting concepts to schema}
\label{sec:lloom_concepts_to_schema_prompt}
\begin{verbatim}
synthesize_schema_from_concepts = '''
Consider the task of a user writing a research paper 
and creating a table to compare and contrast a set of
related papers. Given a list of concepts obtained 
from a set of papers, your task is to create the schema
for this table. \
[List of Concepts] {concepts} \
The goal of the schema is to answer the specific user 
goal when creating the table as follows: \
[User Goal] {user_goal} \
You should find the appropriate number of attributes 
from these concepts that are most useful for comparing
the papers in order to achieve the user goal. Each 
attribute you select to be a part of the schema is 
like a topic covered in the  Related Work section of
the user's paper. \
Return the schema as a JSON object in the following 
format: \"""json {"<attribute 1>": {"definition":
<your definition of why this attribute should be an
axis of comparison>, "output_format":<describe the 
range of output values that will be filled in, is it
numbers, string values or another format>} , ...} """ \
'''
\end{verbatim}

\section{Sequential Prompting Details}
\label{sec:appendix_sequential_prompting_details}
\normalsize
\subsection{Sequential Prompting Pipeline}
Following \citet{wang2025can}, we create a pipeline that sequentially uses information from the full-text of the research papers to iteratively update a candidate schema:
\begin{itemize}
    \item The first step involves summarizing concepts from research papers, similar to the first step of Lloom (\S\ref{sec:appendix_lloom_details}). We reuse the prompt in \S\ref{sec:lloom_summarize_prompts}. The full text of each paper is summarized into a concise yet informative paragraph by concatenating the bullet points. 
    \item We initialize an empty JSON as the schema to begin the process. 
    \item In each iteration, we sample paragraphs summarizing papers in batches of $4$ and prompt the model to update the schema based on this set of papers.\footnote{Unlike \cite{wang2025can}, we do not consider paper selection as part of the task, so we omit that step from their prompting pipeline.} The prompt for this step is provided in \S\ref{sec:prompt_seq_prompting}. The prompt explicitly includes instructions to make add, edit or remove operations on the existing schema. The expectation is that the edit from the current batch of papers also remains faithful to the set of papers seen before by also providing the titles and abstracts of all papers in each step. 
    \item We repeat for 5 sets of iterations through all papers with the order randomized between iterations. 
\end{itemize}

The results reported in \S\ref{tab:generation_results} confirm that this method results in high recall schemas with slightly lower precision than jointly prompting the model with information from all papers at once. 

\subsection{Prompt to update schema from a new batch of papers}
\label{sec:prompt_seq_prompting}
\tiny
\begin{verbatim}
UPDATE_SCHEMA_FROM_NEXT_PAPER = '''
Imagine you are a co-author of a scientific paper and
the first author is creating a table for comparing 
different papers/methods. You are aware of the intent
of the author about the information they want to 
convey via the table. \
You are considering papers one batch at a time and updating
the schema every time you get a new paper. \
Given the current schema, the original author intent 
for the table, the information from the batch of papers, and 
titles+abstracts of all papers, update the schema accordingly.
To update the schema, you can add or remove columns 
as well as modify existing columns as appropriate. You
can also do nothing if the existing schema is good. If
the current schema is empty, create one from scratch. 
Every time you update the schema, make sure you cover 
all relevant information from all papers so far, and 
keep the schema readable and meaningful. \
[Intent] {intent} \
[Current Schema] {curr_schema} \
[New Batch] {new_batch} \
[Summaries Of All Papers] {past_papers} \
Return a JSON object in the following format: \"""json 
{{"<attribute 1>": {{"definition":<your definition of 
why this attribute should be an axis of comparison>, 
"output_format":<describe the range of output values 
that will be filled in, is it numbers, string values 
or another format>}} , ...}} """ \
'''
\end{verbatim}

\section{Fine-tuning open-weight models for schema generation}
\label{sec:appendix_training_details}
\normalsize
We train the Qwen-2.5-3B-Instruct and Llama-3.2-3B-Instruct for schema generation using the following setup. First, we create a training dataset from the ``medium quality'' examples from \citet{newman2024arxivdigestables} (\silverdatasetname) where the input of each example consists of the table intent as well as the concatenated paper titles and abstracts being compared. These tables are not manually checked for parsing errors, are filtered less stringently, and do not have linked full-texts. The output is the JSON schema to be generated. We create a train split ($21168$ examples) and held out validation split ($1115$ examples) and apply the Huggingface \href{https://huggingface.co/docs/transformers/main/en/chat_templating}{Chat Template} to this data. 
The system prompt for the chat template is the same as that of the prompting experiments, provided in \S\ref{sec:prompt_system}. We fine-tune models with traditional supervised fine-tuning, reducing the cross entropy loss on output tokens. We train for $4$ epochs with batch size $1$ and perform sweeps for the learning rate between $1e-06$ and $1e-04$, selecting the best checkpoint using validation loss. Our training happens on Nvidia A100 GPUs. 
At inference time, we use standard HuggingFace \href{https://huggingface.co/docs/transformers/v4.51.3/en/main_classes/pipelines#transformers.Pipeline}{pipelines} after tokenizing examples with the chat template as during training. We sample one output setting the temperature to $0.7$ 
and nucleus sampling \texttt{top\_p} as $0.9$.

\section{Training models for unguided editing of schemas}
\label{sec:appendix_unguided_editing}
We train the Qwen-2.5-3B-Instruct and Llama-3.2-3B-Instruct for unguided schema editing using the following setup. First, we sample output for each example in \silverdatasetname using the GPT-4o prompting (T+A+TI) baseline in \S\ref{sec:joint_prompting_setup}. We create a training dataset for unguided editing where the input of each example consists of the aforementioned generated schema, table intent, as well as the concatenated paper titles and abstracts being compared. The output is the reference JSON schema. We want the model to learn the edit from the generated candidate to the reference. We create a train split ($21168$ examples) and held-out validation split ($1115$ examples) and apply the Huggingface \href{https://huggingface.co/docs/transformers/main/en/chat_templating}{Chat Template} to this data. 
The system prompt for the chat template is the same as that of the prompting experiments, provided in \S\ref{sec:prompt_system}. We fine-tune models with traditional supervised fine-tuning, reducing the cross entropy loss on output tokens. We train for $4$ epochs with batch size $1$ and perform sweeps for the learning rate between $1e-06$ and $1e-04$, selecting the best checkpoint using validation loss. Our training happens on Nvidia A100 GPUs. 
At inference time, we use standard HuggingFace \href{https://huggingface.co/docs/transformers/v4.51.3/en/main_classes/pipelines#transformers.Pipeline}{pipelines} after tokenizing examples with the chat template as during training. We sample one output setting the temperature to $0.7$ 
and nucleus sampling \texttt{top\_p} as $0.9$.

\section{Training heuristics-guided editing models}
\label{sec:appendix_heuristics_guided_editing}
We train the Qwen-2.5-3B-Instruct and Llama-3.2-3B-Instruct for heuristics-guided schema editing using the following setup. 
We want to train models that can add a column (AC) and drop a column (DC) to a candidate schema given the table intent and paper information. 
\begin{itemize}
    \item Add Column (AC): We take the reference schemas from \silverdatasetname and randomly drop one schema item per example. This allows us to create pairs where the output is the reference schema with all schema items, and our 'candidate' schema with one item missing. As a result, when we fine-tune a model to generate the reference as output when given the input of this candidate schema along with the table intent and paper titles and abstracts, we learn the Add Column operation. 
    \item Drop Column (DC): First, we sample output for each example in \silverdatasetname using the GPT-4o prompting (T+A+TI) baseline in \S\ref{sec:joint_prompting_setup}. We then take the reference schemas from \silverdatasetname and augment them with one randomly sampled item from the corresponding GPT-4o generation. This allows us to create pairs where the output is the reference schema with all schema items, and our 'candidate' schema with an extra item that has been added. As a result, when we fine-tune a model to generate the reference as output when given the input of this candidate schema along with the table intent and paper titles and abstracts, we learn the Drop Column operation. 
\end{itemize}

For each operation, we create a train split ($21168$ examples) and held-out validation split ($1115$ examples) in this manner and apply the Huggingface \href{https://huggingface.co/docs/transformers/main/en/chat_templating}{Chat Template} to this data. 
The system prompt for the chat template is the same as that of the prompting experiments, provided in \S\ref{sec:prompt_system}. We fine-tune models with traditional supervised fine-tuning, reducing the cross entropy loss on output tokens. We train for $4$ epochs with batch size $1$ and perform sweeps for the learning rate between $1e-06$ and $1e-04$, selecting the best checkpoint using validation loss. Our training happens on Nvidia A100 GPUs. 
At inference time, we use standard HuggingFace \href{https://huggingface.co/docs/transformers/v4.51.3/en/main_classes/pipelines#transformers.Pipeline}{pipelines} after tokenizing examples with the chat template as during training. We sample one output setting the temperature to $0.7$ 
and nucleus sampling \texttt{top\_p} as $0.9$.

\section{Critique-Guided Editing Details}
\label{sec:appendix_prompting_critiques_details}
\begin{figure*}
    \centering
    \includegraphics[width=\linewidth]{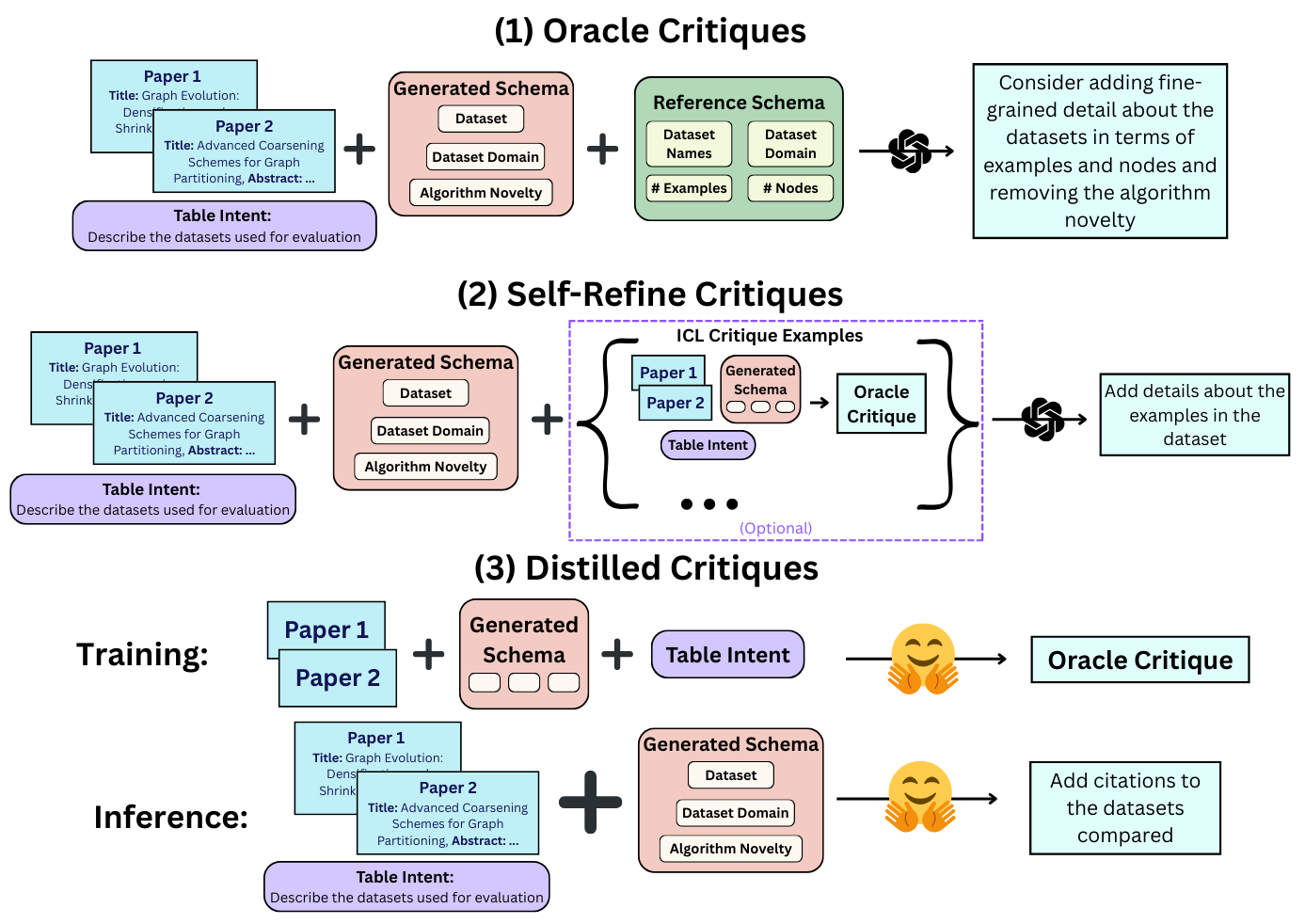}
    \caption{Methods for generating critiques for \S\ref{sec:critiques_editing}.}
    \label{fig:critiques}
    \vspace{-0.5cm}
\end{figure*}
\normalsize
\subsection{Prompting and fine-tuning models for generating critiques}
Here we provide the details for critique-based experiments in \S\ref{sec:critiques_editing}. The model we use to create critiques is \texttt{gpt-4o-2024-08-06} accessed between August 2024 and May 2025. We access the model via the API with the system prompt provided in \S\ref{sec:prompt_system}. We sample one output per example with temperature $0.7$. The different methods we use for critique-generation are shown in \Cref{fig:critiques}.

\begin{itemize}
    \item To create the oracle critiques of the generated schema given the reference, the table intents and paper information, we use prompt \S\ref{sec:prompt_oracle_critique}. We then implement this critique using the prompt in \S\ref{sec:prompt_update_from_critique} to obtain the results in \S\ref{tab:editing_results}.
    \item For the two self-refine baselines, we generate critiques using the table intent and paper information (\S\ref{sec:prompt_generate_critique}) and also by providing ICL examples of critiques concatenated together (\S\ref{sec:prompt_icl_critiques}). We select ICL examples by calculating the $5$ most similar examples in \silverdatasetname to the test example, where similarity is calculated using BERTScore on table intents. We then generate oracle critiques (with \S\ref{sec:prompt_oracle_critique} and concatenate these to create the prompt. We then update the schema based on the critiques generated using the prompt in \S\ref{sec:prompt_update_from_critique} to obtain the results reported in \S\ref{tab:editing_results}. 
    \item For the distilled critiques, we obtain oracle critiques for each example in \silverdatasetname and split these randomly to create a training dataset of $21168$ training examples and $1115$ validating examples. We then fine-tune Llama-3.2-3B-Instruct, Llama-3.2-1B-Base and Qwen-2.5-3B-Instruct. The input consists of the table intent, paper titles and abstracts and the schema generated and the expected output is the oracle critique.
    For the instruct models, we use the Chat Template on the input along with the system prompt in \S\ref{sec:prompt_system}, and for base models, we simply provide the input. We fine-tune the models to reduce the cross-entropy tokens in the output critique, sweeping learning rates from $1e-4$ to $1e-6$ and selecting the best checkpoint using validation loss. We then run inference using the same input format on \datasetname and \harddatasetname to obtain the distilled critiques. We implement the distilled critiques using GPT-4o with the prompt provided in \S\ref{sec:prompt_update_from_critique} to obtain the results in \S\ref{tab:editing_results}.
\end{itemize}

\subsection{Prompts used}
\subsubsection{Prompt for generating a critique to a schema with respect to the table intent}
\label{sec:prompt_generate_critique}
\tiny
\begin{verbatim}
SINGLE_CRITIQUE_SCHEMA_TO_GOAL = '''
Imagine you are a co-author of a scientific paper 
and the first author is creating a table for 
comparing different papers/methods. You are aware
of the intent of the author about the information
they want to convey via the table. \
Given the author intent and the schema of the 
table with the papers being compared, you are 
giving them A SINGLE PIECE OF feedback on the 
schema of the table based on various criteria. \
1. You want the schema to be relevant to the user
goal. Are all the schema items relevant to the 
goal? Is there any information missing in the 
paper information which is relevant to the user
goal which is not included in the schema? \
2. You want the schemas to be non-redundant. If
there is some information shared in the different
aspects of the schema, there is redundancy. If 
there is high redundancy, a user might find it 
hard to understand the table. Are there 
redundant schema items? \
3. You want the schema to be readable in that
the amount of complexity in each of the columns
of the table to be roughly uniform. Are there
some columns that are on different complexity
levels to others which makes the schema less 
readable? \
4. You want the schema to generally be 
informative to the reader. Would the user find
the table informative? Is there any missing 
information in the schema? \
5. You want the schema to be highly specific 
to the user goal. Can you improve the 
specificity somehow? \
6. Are there other dimensions that you notice
as issues with the schema? \
Remember to be HIGHLY CRITICAL. You want the 
feedback to be actionable and you want to help
them improve their work. You are only allowed
to give one SINGLE piece of feedback so make 
sure that you choose the most important piece
of feedback and provide that. 
[Intent] {intent} \
[Schema] {schema} \
[Paper Information] {papers}
Return the feedback in the following JSON 
format: {{'model_feedback':<justification>}} \
'''
\end{verbatim}

\subsubsection{Prompt for updating a schema based on generated critique}
\label{sec:prompt_update_from_critique}
\tiny
\begin{verbatim}
UPDATE_SCHEMA_FROM_SINGLE_CRITIQUE = '''
Imagine you are a co-author of a scientific 
paper and the first author is creating a table
for comparing different papers/methods. You 
are aware of the intent of the author about the
information they want to convey via the table. \
Your advisor has given you feedback on the 
schema of this table that is to be incorporated
to improve the table. \
Given the original schema, the feedback to be 
incorporated, the original author intent for 
the table and information about the papers 
being compared, update the schema accordingly.
If the feedback tells you to drop irrelevant 
columns then remove them from the schema, or 
if the feedback mentions to combine redundant
columns then remove the original ones after 
you combine them, or if the schema asks you 
to be more specific then update the column 
name in the schema, and so on. Your goal is
to incorporate all the feedback and improve 
on the original schema: \
[Intent] {intent} \
[Original Schema] {org_schema} \
[Feedback to be used] {feedback} \
[Compared Papers] {papers} \
Return a JSON object in the following format:
\"""json {{"<attribute 1>": {{"definition":
<your definition of why this attribute should
be an axis of comparison>, "output_format":
<describe the range of output values that will
be filled in, is it numbers, string values or
another format>}} , ...}} """ \
'''
\end{verbatim}

\subsubsection{Prompt for generating an oracle critique based on the reference schema and table intent}
\label{sec:prompt_oracle_critique}
\tiny
\begin{verbatim}
LLM_SINGLE_CRITIQUE_TO_REFERENCE = '''
Imagine you are a teacher and you want to teach
your student how to write a related work table
that compares different papers. You gave them 
an assignment of creating a table to compare a
set of papers. You have the schema of the table
they created as well as a reference table and
information about the papers being compared.  \
You are giving them feedback on the schema they
created. But you also don't want to give them 
the answer. Generate a single critique. Here are
some example issues which you might want to 
critique in the created schema: \
1. You want the schema to match the reference. 
Are there items in the created schema that are 
not present in the reference, are there 
reference items which are missing in the 
created schema? Remember to NOT MENTION THE 
REFERENCE. The student does not know the 
reference exists. You need to frame each 
critique with respect to the paper information.
Items in the reference are important from the
papers, so make sure that you discuss missing
items in that way. Similarly extra items in
the created schema not in the reference are
relatively unimportant in comparing the papers. \
2. You want the schemas to be non-redundant. 
If there is some information shared in the 
different aspects of the schema, there is 
redundancy. If there is high redundancy, a user
might find it hard to understand the table. 
Are there redundant schema items? \
3. You want the schema to be readable in that
the amount of complexity in each of the columns
of the table to be roughly uniform. Are there 
some columns that are on different complexity 
levels to others which makes the schema less 
readable? \
4. You want the schema to generally be 
informative to the reader. Would the user find
the table informative? \
5. Are there other dimensions that you notice 
as issues with the schema? \
Remember to be HIGHLY CRITICAL. You want the
feedback to be actionable, and you do not want
to leak the reference so make sure that you 
always frame your critique in terms of the 
information from the papers and not the 
reference. \
Most importantly you are only allowed to 
suggest a SINGLE edit to the schema, so 
identify the most important issue wrong with
the schema and critique ONLY that one. \
[Created Schema] {gen_schema} \
[Reference Schema] {ref_schema} \
[Paper Information] {papers} \
Return the feedback in the following JSON 
format: {{'critique':<your critique>}} \
'''
\end{verbatim}

\subsubsection{Prompt for generating critique to a schema with respect to table intent and similar examples of critiques}
\label{sec:prompt_icl_critiques}
\tiny
\begin{verbatim}
LLM_SINGLE_CRITIQUE_FROM_SIMILAR = '''
Imagine you are a teacher and you want to teach
your student how to write a related work table
that compares different papers. You gave them 
an assignment of creating a table to compare a
set of papers. You have the schema of the table
they created, the intent that they wanted to 
convey with the table and information about 
the papers being compared.  \
You are giving them feedback on the schema 
they created. But you also don't want to give
them the answer. Generate a single critique.
Here are some example critiques to use as a 
reference to understand the task of critique 
writing: \
[Reference examples of critiques] 
{critique_text} \
Remember to be HIGHLY CRITICAL. You want the 
feedback to be actionable so make sure that 
you always frame your critique in terms of 
the information from the papers. \
Most importantly you are only allowed to 
suggest a SINGLE edit to the schema, so 
identify the most important issue wrong with
the schema and critique ONLY that one. \
[Intent] {intent} \
[Created Schema] {org_schema} \
[Paper Information] {papers} \
Return the feedback in the following JSON 
format: {{'critique':<your critique>}} \
'''
\end{verbatim}

\subsection{Settling on a format of the critique}
\label{sec:appendix_critiques_unused}
\normalsize
We experimented extensively with the format of the natural language critique to be used for the experiments in \S\ref{sec:critiques_editing}. These included:
\begin{itemize}
    \item Detailed paragrah-level critiques that listed out all pieces of feedback to be edited in the generated schema
    \item A series of edits in the form of Add/Remove/Modify operations to be made to the generated schema items
    \item A set of general-purpose rules to be applied to make edits to the schema
    \item Inferring natural language critiques using an LLM ( GPT-4o, o1-mini and Deepseek R1) to identify patterns of errors between the generated and reference schemas 
\end{itemize}
All of these different approaches obtained lower performance than the format we report in the paper, which prompts the model to obtain a single atomic edit to be made to the schema, which corrects the largest error in the schema. The larger the edit recommended by the critique, the higher the oracle performance but the lower was the performance on the test examples. Our experiments were extensive but not exhaustive, so we provide this additional documentation to further research into methods for appropriately critiquing the schema generation task. 


\section{Statistical significance of schema generation results}
\label{sec:appendix_significance_generation}
To test for the significance of the results obtained in \S\ref{tab:generation_results}, we calculate the Recall, Precision, and F1 AUC of each example from different methods and conduct a two-tailed t-test to see if the mean AUC values are different with significance at the 5\% level. Since it is difficult to test for the significance between every baseline in \S\ref{sec:generating_schemas}, we report a subset of results for all the main takeaways in \S\ref{tab:appendix_generation_results_significance}. 

\begin{itemize}
    \item Incorporating table intents into the prompting pipeline increases the mean recall, precision, and F1 AUC for both GPT-4o and Claude-Sonnet with significance (\textbf{T + A + TI} over \textbf{T + A} and \textbf{T + A + C + IR}), confirming their validity as a helpful augmentation to the dataset (\S\ref{sec:motivation_goals}). 
    \item The highest improvement we see from prompting is the \textbf{T + A + TI + ICL} baseline, and we confirm that the mean recall, precision, and F1 AUC are higher than \textbf{T + A + TI} with significance. This demonstrates the value of providing ICL examples to demonstrate the task format (\S\ref{sec:generation_results}).
    \item Sequential prompting with \textbf{T + FT + TI} improves recall AUC but, importantly, has \textbf{lower} precision and F1 AUC as compared to both joint prompting baselines of \textbf{T + A + TI + ICL} and \textbf{T + A + TI}, each with statistical significance (\S\ref{sec:generation_results}). 
    \item Both fine-tuned Qwen-2.5-3B-Instruct and Llama-3.2-3B-Instruct achieve lower recall but higher precision and F1 AUC than GPT-4o on the same input with significance confirming that specialized open-weight models are competitive with black box LLMs. 
\end{itemize}

\begin{table}[]
\resizebox{\columnwidth}{!}{%
\begin{tabular}{@{}rrccc@{}}
\toprule
\textbf{Model} & \textbf{Input} & \textbf{Recall} & \textbf{Precision} & \textbf{F1} \\ \midrule
\multirow{3}{*}{\textbf{GPT-4o}} & \textbf{Baseline - T + A} & 0.1954 & 0.1791 & 0.1806 \\
 & \textbf{Baseline - T + A + C + IR} & 0.2409 & 0.2159 & 0.2214 \\   
 & \textbf{T + A + TI} & 0.2811* & 0.2648* & 0.2666* \\ \cmidrule(l){2-5}
 & \textbf{Baseline - T + A + TI} & 0.2811 & 0.2648 & 0.2666 \\
 & \textbf{T + A + TI + ICL} & 0.2941* & 0.2807* & 0.2811* \\
 \cmidrule(l){2-5}
 & \textbf{Baseline - Joint - T + A + TI} & 0.2811 & 0.2648 & 0.2666 \\
 & \textbf{Sequential - T + FT + TI} & 0.3122* & 0.2297* & 0.2541* \\
 \midrule
\multirow{3}{*}{\textbf{Claude Sonnet}} & \textbf{Baseline - T + A} & 0.2118 & 0.1841 & 0.1903 \\
& \textbf{Baseline - T + A + C + IR} & 0.2554 & 0.2360 & 0.2367 \\ 
 & \textbf{T + A + TI} & 0.2655* & 0.2313* & 0.2407* \\ \cmidrule(l){2-5} 
 & \textbf{Baseline - T + A + TI} & 0.2655 & 0.2313 & 0.2407 \\ 
 & \textbf{T + A + TI + ICL} & 0.2903* & 0.2505* & 0.2617* \\
 \cmidrule(l){2-5} 
  & \textbf{Baseline - Joint - T + A + TI} & 0.2655 & 0.2313 & 0.2407 \\ 
  & \textbf{Sequential - T + FT + TI} & 0.3138* & 0.2090* & 0.2386* \\
 \midrule
 \textbf{GPT-4o} & \textbf{Baseline - T + A + TI} & 0.2811 & 0.2648 & 0.2666 \\
 \textbf{Qwen-3B-Instruct} & \textbf{T + A + TI} & 0.2638* & 0.3143* & 0.2772* \\
 \midrule
 \textbf{GPT-4o} & \textbf{Baseline - T + A + TI} & 0.2811 & 0.2648 & 0.2666 \\
 \textbf{Llama-3B-Instruct} & \textbf{T + A + TI} & 0.2684* & 0.3006* & 0.2747* \\
 \bottomrule
\end{tabular}%
}
\caption{Statistical significance testing of GPT-4o and Claude-Sonnet schema generation results, given varying inputs. Here T stands for paper titles, A for abstracts, TI for table intents, FT for full-texts, C for table captions, IR for in-text references, ICL for in-context examples. Values marked with an asterisk are cells where the mean is different from the baseline row in the same section with significance at the 5\% level, magnitude indicating the direction of difference. See \S\ref{sec:appendix_significance_generation} for discussion.} 
\label{tab:appendix_generation_results_significance}
\end{table}



\section{Qualitative analysis of critiques}
\label{sec:critiques_error_analysis}
\normalsize
To understand why model-generated critiques are not as helpful as the oracle critiques in \S\ref{sec:critiques_editing}, one author of this paper manually examined $25$ examples of critiques to categorize common patterns of failure. These examples were randomly sampled from the Self-Refine critique baseline in \S\ref{sec:critiques_editing}. The most prevalent categories are as follows:
\begin{itemize}
    \item \textbf{Generic feedback:} One common issue ($6$ examples) is when the model provides a broadly applicable critique without being specific to either the table intent or the papers being compared. An example is provided in \S\ref{tab:critique_example_generic} where the critique asks for an edit to introduce a \emph{Task} aspect of comparison, which is not very relevant to the particular candidate schema. This leads the model to make fairly generic and unhelpful edits.
    \item \textbf{Incorrect reasoning:} A challenging issue to automatically detect is one where the model-generated critique might seem correct at first glance but does not hold up to fine-grained scrutiny ($5$ examples). Consider \S\ref{tab:critique_example_reasoning}, where the critique points out a potential overlap between two schema items and recommends that these be combined into one aspect. However the two aspects being spoken about speak about the deployment in various applications and open issues, which are fairly broad to be combined into one aspect. 
    \item \textbf{Factual Errors:} The third pattern of errors was due to low level factual errors in the critique ($5$ examples). In \S\ref{tab:critique_example_factual}, the model-generated critique asks for an edit that would introduce an element about the challenges in using smart sensors in building management systems. This is largely covered within the aspect ``Challenges and Limitations in Deployment'', that explicitly mentions smart sensors in the definition. Training LLMs to avoid these hallucinations would lead to more high quality critiques. 
\end{itemize}

\begin{table*}[]
\resizebox{\textwidth}{!}{%
\centering
\begin{tabular}{@{}L{3cm}C{12cm}L{3cm}@{}}
\toprule
Table Intent & Generated Schema & Critique \\ \midrule
How do the datasets included in the GAIA Search tool compare in terms of size and availability, and what implications does this have for their use in large-scale language model training? & \begin{tabular}[c]{@{}L{12cm}@{}}\{\\     "Dataset Size": \{\\         "definition": "The size of a dataset is a crucial factor for training large-scale language models, as larger datasets can potentially provide more diverse and representative samples, leading to better model generalization and performance.",\\         "output\_format": "Numeric value representing the size of the dataset in gigabytes (GB) or terabytes (TB)."\\     \},\\     "Data Diversity": \{\\         "definition": "The diversity of the data in a dataset can impact how well a language model trained on it can generalize across different domains and tasks. A diverse dataset contains text from various sources and domains.",\\         "output\_format": "Descriptive string indicating the range of data sources and domains included in the dataset."\\     \},\\     "Language Coverage": \{\\         "definition": "Language coverage refers to the number and variety of languages represented in a dataset. Multilingual datasets have the potential to train models that can understand and generate text in multiple languages.",\\         "output\_format": "String describing the number of languages and any specific language focus (e.g., monolingual, multilingual with specific languages)."\\     \},\\     "Data Availability": \{\\         "definition": "The availability of a dataset determines how accessible it is to the research community, affecting who can use it for model training and experimentation.",\\         "output\_format": "String indicating whether the dataset is publicly available, along with any restrictions or accessibility notes."\\     \},\\     "Ethical Considerations": \{\\         "definition": "Ethical considerations involve the potential issues related to the content of the dataset, such as biases, harmful content, and governance, which can impact the fairness and safety of models trained on the data.",\\         "output\_format": "Descriptive string outlining any ethical considerations or measures taken to address them in the dataset."\}\}\end{tabular} & The schema you've created is quite comprehensive in terms of dataset characteristics and ethical considerations, but it misses an important dimension that could significantly aid in understanding the datasets: the 'Task' column. This column is crucial because it specifies the type of machine learning tasks that the datasets are intended to support, such as classification, translation, or grounding. \\
\bottomrule
\end{tabular}%
}
\caption{Example of a critique that provides generic feedback that is not very actionable}
\label{tab:critique_example_generic}
\end{table*}

\begin{table*}[]
\resizebox{\textwidth}{!}{%
\centering
\begin{tabular}{@{}C{3cm}C{12cm}C{3cm}@{}}
\toprule
Table Intent & Generated Schema & Critique \\ \midrule
What are the different aspects of Mobile Edge Computing (MEC) performance explored in recent academic surveys, and how do these aspects contribute to the understanding and improvement of MEC systems? & \begin{tabular}[c]{@{}L{12cm}@{}}\{\\     "Latency and Reliability": \{\\         "definition": "Latency and reliability are critical performance metrics for MEC systems, especially in applications like virtual reality and vehicle-to-everything (V2X), where quick response times and dependable service are necessary. These aspects are essential for understanding how MEC can meet the demands of real-time and mission-critical applications.",\\         "output\_format": "string values, such as 'low latency', 'high reliability', 'ultra-reliable', 'mission-critical'"\\     \},\\     "Network Infrastructure and Architecture": \{\\         "definition": "The design and architecture of the network infrastructure are fundamental to the successful deployment of MEC systems. This attribute encompasses the distribution of computing resources, proximity to end devices, and integration with existing network systems, which all affect the performance and scalability of MEC.",\\         "output\_format": "string values, such as 'distributed', 'centralized', 'integrated with 5G', 'small cell base stations'"\\     \},\\     "Applications and Use Cases": \{\\         "definition": "Understanding the applications and use cases for MEC is vital to grasp its potential and limitations. Different applications have varying requirements, such as those in virtual reality or vehicular networks, which drive the need for specific MEC features and enhancements.",\\         "output\_format": "list of string values, such as {[}'virtual reality', 'V2X', 'edge AI', 'vehicular applications'{]}"\\     \},\\     "Challenges and Open Issues": \{\\         "definition": "Identifying and addressing the challenges and open issues in MEC systems is crucial for their improvement. This includes aspects like scalability, security, and resource allocation, which need to be tackled to ensure efficient and effective MEC deployment.",\\         "output\_format": "list of string values, such as {[}'scalability', 'security', 'resource allocation', 'response time'{]}"\\     \},\\     "Future Research Directions": \{\\         "definition": "Exploring future research directions helps in setting a roadmap for the evolution of MEC technologies. It involves investigating emerging trends, technological advancements, and potential innovations that can enhance MEC performance.",\\         "output\_format": "list of string values, such as {[}'5G integration', 'AI at the edge', 'enhanced VEC applications'{]}"\\     \}\\ \}\end{tabular} & The schema currently includes 'Applications and Use Cases' and 'Challenges and Open Issues,' which overlap in scope, as both address aspects of MEC performance and challenges. Consider consolidating these columns into one that distinctly focuses on performance characteristics, as these are central themes in the papers. This will reduce redundancy and improve the clarity and informativeness of your comparison table. \\ 
\bottomrule
\end{tabular}%
}
\caption{Example of a critique with incorrect reasoning about the overlap between the two aspects, one which discusses weaknesses and the other discusses examples deployment.}
\label{tab:critique_example_reasoning}
\end{table*}

\begin{table*}[]
\resizebox{\textwidth}{!}{%
\begin{tabular}{@{}C{3cm}C{12cm}C{3cm}@{}}
\toprule
Table Intent & Generated Schema & Critique \\ \midrule
What are the existing research trends and methodologies in Intelligent Energy Management Systems (IEMS) for buildings, and how does the current study address the gaps or limitations identified in these previous works? & \begin{tabular}[c]{@{}L{12cm}@{}}\{\\     "Recommender Systems in Energy Management": \{\\         "definition": "Recommender systems are increasingly integrated into energy management solutions for buildings, leveraging IoT and AI technologies to enhance energy efficiency. These systems promote energy-saving behavior and reduce carbon emissions, and this attribute is crucial to understanding their role and development in IEMS.",\\         "output\_format": "string values describing the presence, type, and role of recommender systems in the study"\\     \},\\     "Modeling Approaches for Energy and Comfort Optimization": \{\\         "definition": "Different modeling approaches, such as white box, black box, and gray box models, are used to optimize building energy management and occupant comfort. Understanding these models' strengths and weaknesses helps evaluate their effectiveness in intelligent energy management systems.",\\         "output\_format": "string values categorizing the types of models used and their advantages/disadvantages"\\     \},\\     "Challenges and Limitations in Deployment": \{\\         "definition": "Identifying the challenges and limitations faced in the deployment of smart sensors in intelligent energy management systems in buildings",\\         "output\_format": "string values describing identified challenges and limitations"\\     \},\\     "Evaluation Metrics and Incentive Measures": \{\\         "definition": "Evaluation metrics and incentive measures are used to assess the performance of IEMS and encourage their adoption. This attribute is necessary to compare how different studies measure success and promote system implementation.",\\         "output\_format": "string values detailing the evaluation metrics and incentive measures discussed"\\     \},\\     "Technological Integration and Advancement": \{\\         "definition": "The integration of advanced technologies, such as IoT and AI, into energy management systems is crucial for improving their functionality and applicability. This attribute outlines how technological advancements are harnessed in the studies to achieve energy efficiency.",\\         "output\_format": "string values explaining the technological components and their integration"\\     \}\\ \}\end{tabular} & The current schema does not explicitly address the specific challenges or limitations related to specific technological components like smart sensors and actuators, which are mentioned in the papers as crucial for effective system deployment. Including a column specifically focused on 'Challenges and Limitations Related to Smart Sensors and Actuators' would provide a more comprehensive view of the technical obstacles in deploying intelligent systems in buildings. \\ \bottomrule
\end{tabular}%
}
\caption{Example of a critique with a factual error as the ``Challenges in Deployment'' aspect covers the concern mentioned in the critique about smart sensors}
\label{tab:critique_example_factual}
\end{table*}

\section{Use of AI assistants}
The authors acknowledge that AI assistants were used to refine parts of the writing of this work in terms of framing, wording and coherence of different sections. 

\end{document}